%% file: main.tex
\pdfoutput=1
\documentclass[11pt]{article}

\usepackage[preprint]{acl}

\usepackage{times}
\usepackage{latexsym}
\usepackage{graphicx}
\usepackage{amssymb}

\usepackage[T1]{fontenc}
\usepackage{booktabs}
\usepackage{colortbl}
\usepackage{xcolor}
\usepackage{pgfmath}
\usepackage{multirow}
\usepackage{subcaption}
\usepackage{amsmath}
\usepackage{tikz}
\usepackage{hyperref}
\usepackage{xurl}
\usepackage{lscape}
\usepackage{url}
\usepackage{array}
\usepackage{longtable}

\usepackage[utf8]{inputenc}

\usepackage{microtype}

\usepackage{inconsolata}

\input{defs}

\title{MILU: A Multi-task Indic Language Understanding Benchmark}

\author{
 \textbf{Sshubam Verma\textsuperscript{1}} \quad
 \textbf{Mohammed Safi Ur Rahman Khan\textsuperscript{1,2}} 
 \\
 \textbf{Vishwajeet Kumar\textsuperscript{3}} \quad
 \textbf{Rudra Murthy\textsuperscript{3}} \quad
 \textbf{Jaydeep Sen\textsuperscript{3}}
\\
 \textsuperscript{1}Nilekani Centre at AI4Bharat \quad
 \textsuperscript{2}Indian Institute of Technology, Madras \\
 \textsuperscript{3}IBM Research, India
\\
  \small{
   \textbf{Correspondence:} \texttt{\{sshubamverma, safikhan\}@ai4bharat.org, \{vishk024, rmurthyv\}@in.ibm.com}
 }
 \\
 \includegraphics[height=1.1em]{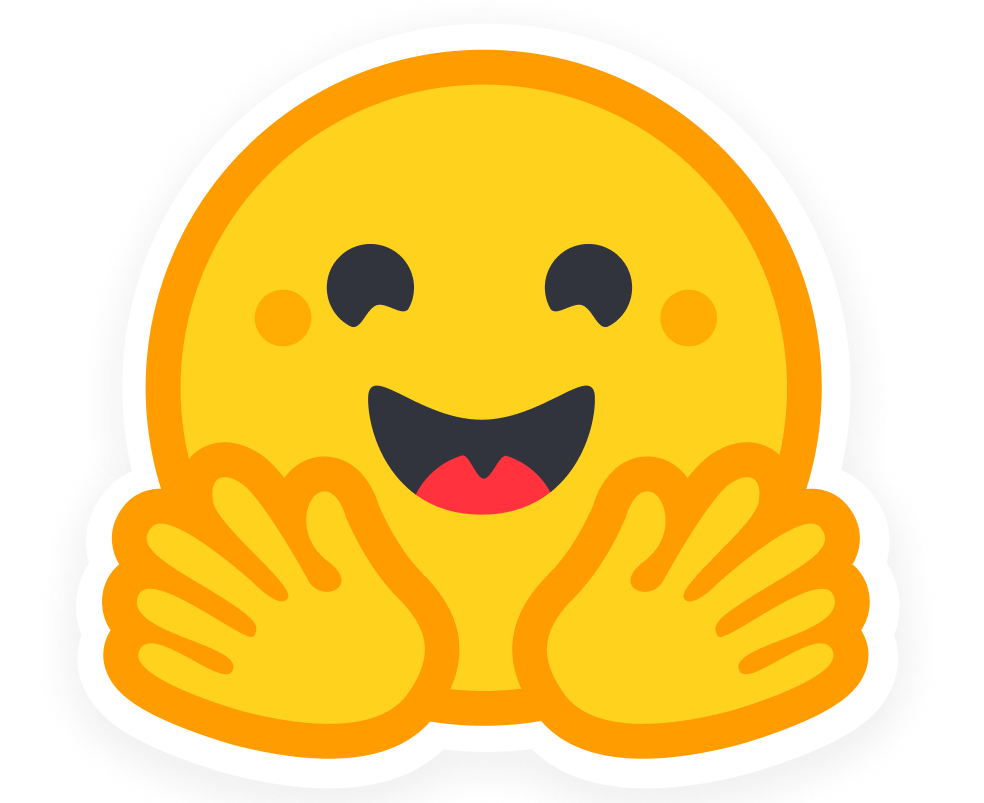}~\small{\url{https://huggingface.co/datasets/ai4bharat/MILU}}
 \\
 \includegraphics[height=1em]{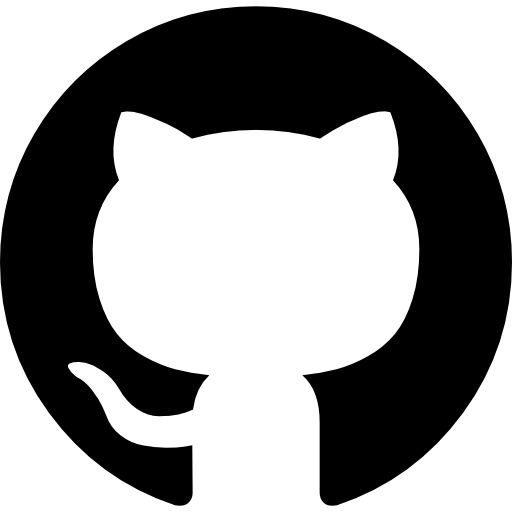}~\small{\url{https://github.com/AI4Bharat/MILU}}
}

\begin{document}
\maketitle

\input{Sections/0_abstract}
\input{Sections/1_introduction}
\input{Sections/2_related}

\input{Sections/3_benchmark}

\input{Sections/4_experiments}

\input{Sections/5_results}

\input{Sections/conclusion}
\input{Sections/limits}
\input{Sections/acknowledgments}
\input{Sections/ethics}

\bibliography{main}

\appendix
\input{Sections/appendix}

\end{document}

%% file: defs.tex
\newcommand{\milu}{\textsc{MILU}}
\newcommand{\indictrans}{\textsc{IndicTrans2}}
\newcommand{\nvembed}{\textsc{NV-Embed-v2}}
\newcommand{\gpt}{\textsc{GPT-4o}}
\newcommand{\gptmini}{\textsc{GPT-4o-mini}}

\definecolor{mygreen}{HTML}{02a61b}
\definecolor{myred}{HTML}{cc0000}

\definecolor{pert_green}{HTML}{02a61b}
\definecolor{pert_red}{HTML}{cc0000}

\definecolor{my_orange}{HTML}{FF7F0F}
\definecolor{my_green}{HTML}{2AA02B}
\definecolor{my_blue}{HTML}{1F77B4}

\definecolor{lightorange}{RGB}{255,235,205}
\definecolor{darkorange}{RGB}{255,140,0}

%% file: Sections/0_abstract.tex
\begin{abstract}

Evaluating Large Language Models (LLMs) in low-resource and linguistically diverse languages remains a significant challenge in NLP, particularly for languages using non-Latin scripts like those spoken in India. Existing benchmarks predominantly focus on English, leaving substantial gaps in assessing LLM capabilities in these languages. We introduce \milu{}—\textbf{M}ulti-task \textbf{I}ndic \textbf{L}anguage \textbf{U}nderstanding Benchmark—a comprehensive evaluation benchmark designed to address this gap. \milu{} spans 8 domains and 41 subjects across 11 Indic languages, reflecting both general and culturally specific knowledge. With an India-centric design, \milu{} incorporates material from regional and state-level examinations, covering topics such as local history, arts, festivals, and laws, alongside standard subjects like science. We evaluate over 42 LLMs, and find that current LLMs struggle with \milu{}, with GPT-4o achieving the highest average accuracy at 74\%. Open multilingual models outperform language-specific fine-tuned models, which perform only slightly better than random baselines. Models also perform better in high-resource languages as compared to low-resource ones. Domain-wise analysis indicates that models perform poorly in culturally relevant areas like Arts \& Humanities and Law \& Governance compared to general fields like STEM. To the best of our knowledge, \milu{} is the first of its kind benchmark focused on Indic languages, serving as a crucial step towards comprehensive cultural evaluation. All code, benchmarks, and artifacts will be made publicly available to foster open research.

\end{abstract}

%% file: Sections/1_introduction.tex
\section{Introduction}

Recent advancements in Large Language Models (LLMs) have reshaped the field of NLP, by enabling these models to perform a variety of tasks across diverse domains~\cite{doddapaneni-etal-2023-towards,openai2023gpt4, geminiteam2024geminifamilyhighlycapable,claude}. While many LLMs now claim to support multiple languages, there is still a huge discrepancy in their performance in English and other languages~\cite{liu2024omgeval}. Particularly languages using non-Latin scripts, such as those in India, are affected the most by this discrepancy~\cite{ahuja2023mega}. One key reason for this is the absence of high-quality benchmarks for these languages. Well-designed benchmarks are crucial in driving model development by revealing limitations and guiding improvements~\cite{mcintosh2024inadequacies}. However, most existing benchmarks focus primarily on English, \textit{leaving significant gaps in evaluating LLMs capability in low-resource and linguistically diverse languages.}

\begin{figure}
    \centering
    \includegraphics[width=\columnwidth]{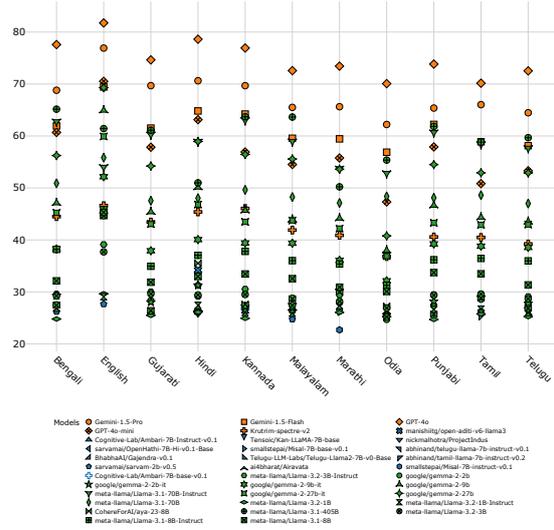}
    \caption{Average performance of all the evaluated models on \milu. The closed models are shown in \textcolor{my_orange}{Orange}, the open models are shown in \textcolor{my_green}{Green}, and the language-specific models are shown in \textcolor{my_blue}{Blue}.}
    \label{fig:intro}
\end{figure}

India’s diversity presents a unique challenge for these models. With over 1.4 billion people speaking more than 120 languages and around 19,500 dialects across 28 states \cite{javed-etal-2024-indicvoices}, many of which are underrepresented in NLP research, the need for not just linguistic but culturally appropriate benchmarks becomes urgent~\cite{doddapaneni-etal-2023-towards,indicgenbench}. Standard benchmarks like \textsc{MMLU}~\cite{mmlu} and \textsc{AGIEval}~\cite{zhong2023agieval}, while useful for evaluating general world knowledge, falls short in capturing these intricacies of India. Each state in India has its own history, traditions, festivals, and art forms, forming a rich cultural mosaic. Translating existing English benchmarks into Indian languages fails to capture this knowledge, which is required for real-world applications. Given the increasing deployment of LLMs in tasks that directly impact local populations, \textit{the need for a benchmark that evaluates both linguistic competence and cultural understanding has become more pressing than ever.}

In this work, we introduce \milu-\textbf{M}ulti-task \textbf{I}ndic \textbf{L}anguage \textbf{U}nderstanding Benchmark, a comprehensive evaluation dataset designed to address these gaps. \milu~spans 8 domains and 41 subjects across 11 Indic languages, reflecting both general and culturally specific knowledge. We designed \milu~with an India-first perspective by collecting questions from various national, state, and regional exams. These questions include culturally relevant subjects such as local history, arts, festivals, and laws, alongside traditional academic subjects like science. Following previous efforts~\cite{mmlu, zhong2023agieval}, we create this benchmark by collecting questions from over 1500 competitive exams from India. We focus on region-specific exams to authentically capture local knowledge in the respective language. 

We evaluate 45 different LLMs - a mix of closed proprietary, open-source, and language-specific models- on \milu. Our findings suggest that models struggle with \milu, with GPT-4o achieving the highest average accuracy at 74\%. Interestingly, open multilingual models outperform language-specific models, which only achieve slightly better than random scores. Our analysis of in-context learning reveals that adding more examples improves performance in base models, but the effect on instruct models remains inconclusive. We also explore how performance scales with the number of parameters, finding significant improvements as model size increases. Our domain-wise analysis reveals that models perform poorly in culturally relevant areas, such as Arts \& Humanities and Social Sciences, compared to more general fields like STEM. All the artifacts will be released publicly.



%% file: Sections/2_related.tex
\section{Related Works}
\noindent \textbf{Large Language Models (LLMs): }
Recent LLMs, both proprietary-GPT-4o, GPT-4o-mini, Claude-3, and Gemini~\cite{geminiteam2024geminifamilyhighlycapable}-and open-source-the Llama-3~\cite{dubey2024llama} and Gemma~\cite{team2024gemma} series-have demonstrated significant improvements across various tasks and benchmarks~\cite{chiang2024chatbot, wang2024mmlupro}. While these models are primarily trained in English, many claim to include a reasonable amount of multilingual data in their pretraining corpus~\cite{team2024gemma, dubey2024llama, aryabumi2024aya}. Additionally, significant progress has been made in developing language-specific models~\cite{ustun2024aya, nguyen2023seallms}, including for Indic languages~\cite{gala2024airavata, balachandran2023tamilllama, kohli2023building}. Most of these models are built on top of stronger English base models by (optionally) either continually pretraining on smaller, language-specific datasets~\cite{zhao2024llama}, or by language-specific instruction fine-tuning~\cite{closerlook}. In this work, we conduct a comprehensive evaluation of the performance of these models on ~\milu.

\noindent \textbf{LLM Evaluation Benchmarks: }
Over the years, various benchmarks have been developed to evaluate the performance of large language models (LLMs). Recent benchmarks such as \textsc{MMLU}~\cite{mmlu}, \textsc{MMLUPro}~\cite{wang2024mmlupro}, \textsc{AGIEval}~\cite{zhong2023agieval}, \textsc{BigGenBench}~\cite{kim2024biggen}, and \textsc{HellaSwag}~\cite{zellers2019hellaswag} assess these models across a wide range of tasks. However, these benchmarks primarily focus on English, and progress on multilingual benchmarks has been comparatively slower \cite{doddapaneni-etal-2023-towards, kumar2022indicnlg}. Some popular multilingual benchmarks include \textsc{OMGEval}~\cite{liu2024omgeval}, \textsc{XTREME}~\cite{hu2020xtreme}, and \textsc{XQuAD}~\cite{artetxe2019crosslingual}, though these are largely limited to simple natural language understanding (NLU) tasks. As a result, previous studies often rely on translations of popular English benchmarks to evaluate performance in other languages~\cite{dubey2024llama, gala2024airavata}. This approach is suboptimal, as it fails to account for cultural nuances and concepts unique to specific languages.

\noindent \textbf{Indic LLM Evaluation Benchmarks: }
The first major Indic language benchmarks, \textsc{IndicGLUE}~\cite{kakwani-etal-2020-indicnlpsuite} and \textsc{IndicNLG-Benchmark}~\cite{kumar2022indicnlg}, cover 11 languages, focusing on various language understanding and generation tasks, respectively. \textsc{IndicXTREME}~\cite{doddapaneni-etal-2023-towards} extends these efforts to include all 22 scheduled Indian languages for NLU evaluation. More recently, \textsc{IndicGenBench}~\cite{indicgenbench} offers a more comprehensive evaluation of multilingual language generation by consolidating multiple generation tasks. In addition, the \textsc{Indic-QA Benchmark}~\cite{singh2024indic} assesses context-based question-answering (QA) performance in 11 Indian languages, while \textsc{L3Cube-IndicQuest}~\cite{rohera2024l3cubeindicquest} explores regional knowledge through a translation-based approach to data creation.
More recently, \textsc{Pariksha}~\cite{watts-etal-2024-pariksha} evaluated more than 25 Indic models and created dynamic leaderboards by collecting more than 90,000 human preferences.
Other initiatives, such as \textsc{Airavata}~\cite{gala2024airavata} and the \textsc{Indic-LLM-Leaderboard}~\cite{indic-llm-leaderboard}, focus on translating existing English benchmarks into Indian languages. 
In contrast, \milu~addresses this limitation by centering on India-specific topics and questions, providing a more culturally relevant and context-aware evaluation.

%% file: Sections/3_benchmark.tex
\section{MILU: The IndicMMLU Benchmark}
In this section, we describe the collection process (\S\ref{sec: questions}), the cleaning and filtering process (\S\ref{sec: cleaning}), and present the analysis (\S\ref{sec: analysis}).

\subsection{Questions Curation}
\label{sec: questions}
\milu~is a large, multi-domain test set containing multiple-choice based questions (MCQs) taken from over 41 subjects with an emphasis on India-specific knowledge. This benchmark covers many domains, including Science, Social Sciences, Humanities, Arts, Business Studies, and Law, among others. \milu~is designed as a culturally relevant benchmark to assess general problem-solving abilities and India-specific knowledge. These questions were sourced following an approach similar to \textsc{AGIEval}~\cite{zhong2023agieval}, collecting the questions from various public exams taken by individuals intending to either pursue higher studies or seek career advancements, such as qualification tests and national and state-level civil services exams, among others.

We gathered exam-specific questions by scraping various online exam portals that offer previously released question papers from various exams in multiple different languages. These portals typically tag questions manually with topic names and language details, and subject experts ensure the accuracy of the answers. Our benchmark includes questions from over 40 different types of exams conducted both at the national and state levels over recent years. Regional state exams are particularly valuable as they cover various state-level topics and emphasize the official language of each state. Further details about the different exams included in our collection are provided in the Appendix \ref{app: exams}.

In total, we collected more than 150K questions across 11 Indian Languages- Bengali (\textit{bn}), Gujarati (\textit{gu}), Hindi (\textit{hi}), Kannada (\textit{kn}), Malayalam (\textit{ml}), Marathi (\textit{mr}), Odia (\textit{or}), Punjabi (\textit{pn}), Tamil (\textit{ta}), Telugu (\textit{te}), and English (\textit{en})-spanning 41 diverse subjects. English questions are also included as these often address Indian culture-specific content, which is notably missing from existing popular benchmarks. Table \ref{tab: mmlu_indicmmlu_comparison} presents a brief comparison of the type of questions present in \textsc{MMLU}~\cite{mmlu} and \milu.

\begingroup
\setlength{\tabcolsep}{3pt} 
\renewcommand{\arraystretch}{1} 
\setlength\fboxsep{3pt}
\begin{table}[t!]
\centering
\scriptsize
\begin{tabular}{p{1.37in} p{1.37in}} 

\midrule
\multicolumn{1}{c}{\textbf{\textsc{MMLU}}} & \multicolumn{1}{c}{\textsc{\textbf{MILU}}} \\ 
\midrule
Which principle was established by the Supreme Court's decision in \textcolor{myred}{Marbury v. Madison}? & Which of the decisions was taken by \textcolor{mygreen}{Maharashtra government} on 24th November 2001 that had a far-reaching impact on the education field? \\
\midrule
At breakfast, lunch, and dinner, \textcolor{myred}{Joe} randomly chooses with equal probabilities either an apple, an orange, or a banana to eat. On a given day, what is the probability that \textcolor{myred}{Joe} will eat at least two different kinds of fruit? & \textcolor{mygreen}{Shyam's} monthly income is Rs. 12,000. He saves Rs. 1200. Find the percent of his savings and his expenditure. \\
\midrule
The \textcolor{myred}{Space Shuttle} orbits 300 km above Earth's surface; Earth's radius is 6,400 km. What is the gravitational acceleration experienced by the Space Shuttle? & \textcolor{mygreen}{FASTag} at the Toll Gates uses which waves? \\
\midrule
The primary goal of the \textcolor{myred}{Gramm-Rudman Acts of 1985 and 1987} was to & Under the \textcolor{mygreen}{Pradhan Mantri Kisan Samman Nidhi Scheme}, the financial help provided to the small and marginal farmers is \_\_\_\_\_\_\_\_\_\_\_ . \\
\midrule
A \textcolor{myred}{guitar} string creates a sound wave of known frequency. Which of the following describes a correct and practical method of measuring the wavelength of the sound wave with a meterstick? & In a \textcolor{mygreen}{Sitar}, which type of sound vibrations are produced? \\
\midrule
\end{tabular}
\caption{Comparison of some question from \textsc{MMLU}~\cite{mmlu} with similar questions from \milu. The \textcolor{pert_red}{Red} highlights the global concepts covered in \textsc{MMLU}, while \textcolor{pert_green}{Green} highlights the India-centric concepts covered in \milu.}
\label{tab: mmlu_indicmmlu_comparison}

\end{table}
\endgroup
 
\subsection{Data Cleaning and Filtering}
\label{sec: cleaning}
Despite our best efforts to maintain the quality of questions collected, some amount of noise or errors may still be present. To address potential noise in the questions, we employ multiple layers of manual and automated cleaning filters. Initially, we manually review a large sample of questions to detect and eliminate potential sources of noise. During the collection process, we exclude any reading-comprehension-style questions, images-based questions, and those with more than four answer options to ensure uniformity and consistency. To remove incorrect language entries, we utilize a combination of \textsc{IndicLID}~\cite{madhani-etal-2023-bhasa} and Unicode-based filtering~\cite{khan2024indicllmsuite}, ensuring that the questions are in the correct language. To further refine the dataset, we remove any duplicate questions to retain only the unique ones. As a final step, we manually verify a sample of questions from each language to ensure accuracy and correct any remaining errors.
\begin{figure}
    \centering
    \includegraphics[width=\columnwidth]{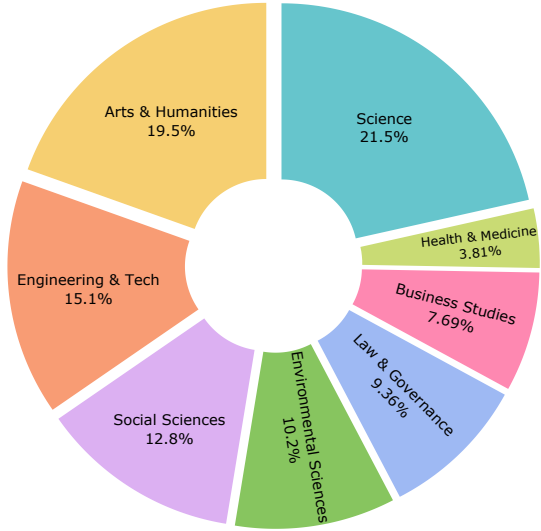}
    \caption{Distribution of the number of questions across different domains, averaged across all languages. Refer to Section (\S\ref{sec: analysis}) for more details.}
    \label{fig:tag}
\end{figure}

Upon examination, we found that approximately 45\% of questions were accurately labeled with a topic name, while the remaining questions lacked this information. To address this issue, we first translate the untagged questions into English using \indictrans~\cite{gala2023indictrans2} and then prompt \gptmini~model to assign an appropriate topic name to the question. Finally, in total, we get around 20K tags. However, these tags are highly fine-grained, often having a heavy overlap. To organize them, we embed the tags using the \nvembed~\cite{lee2024nvembed} model and apply K-means clustering to group tags into 50 clusters. We manually review these clusters and assign appropriate subject labels. Following the manual merging of related clusters, we determine 41 distinct subject names, which fall into eight main domains: Arts and Humanities, Social Sciences, Environmental Sciences, Law and Governance, Health and Medicine, Science, Engineering and Technology, and Business Studies. The final distribution of domains within \milu~is shown in Figure \ref{fig:tag}. Detailed analysis of question distribution by topic \& language is provided in Appendix \ref{app: distrb}.

Finally, we observed that some topics in certain languages had less than 100 questions. To ensure thorough evaluation across all subjects and languages, we aimed to have at least 100 questions per subject in each language. For subjects with insufficient questions, we sampled questions from the English set from that subject and translated them into the required language using \gpt. We chose \gpt~over specialized translation models for their ability to remain task-aware during translation \cite{ahuja2024sphinx}, ensuring the translated content aligns with the intent of the question. 

In total, we release around 79K questions across 41 subjects across 8 domains in 11 languages, capping each subject-language pair at 500 questions for feasible evaluations.

\begingroup
\setlength{\tabcolsep}{6pt} 
\renewcommand{\arraystretch}{1}

\begin{table}[]
\centering
\footnotesize
\begin{tabular}{l|ccc}
\toprule
\textit{\textbf{lang}} & \textbf{\begin{tabular}[c]{@{}c@{}}Total Qs\end{tabular}} & \textbf{\begin{tabular}[c]{@{}c@{}}Total \\ Translated Qs\end{tabular}} & \textbf{\begin{tabular}[c]{@{}c@{}}Avg Words \\ Per Q\end{tabular}} \\
\midrule
\textit{bn}       & 6638 & 1601 & 15.12        \\
\textit{gu}       & 4827 & 2755 & 16.12        \\
\textit{hi}       & 14837 & 115  & 20.61        \\
\textit{kn}       & 6234 & 1522 & 12.42        \\
\textit{ml}       & 4321 & 3354 & 12.39        \\
\textit{mr}       & 6924 & 1235 & 18.76         \\
\textit{or}       & 4525 & 3100 & 14.96       \\
\textit{pa}       & 4099 & 3411 & 19.26         \\
\textit{ta}       & 6372 & 1524 & 13.14        \\
\textit{te}       & 7304 & 1298 & 15.71        \\
\textit{en}       & 13536         & -    & 22.07        \\
\midrule
\textit{\textbf{total}}  & \textbf{79617}         & \textbf{19915}           & \textbf{16.77}       \\
\bottomrule
\end{tabular}
\caption{Overall statistics of \milu. Refer to Section (\S\ref{sec: analysis}) for more details.}
\label{tab: milu_stats}
\end{table}
\endgroup

\subsection{Data Analysis}
\label{sec: analysis}
Table \ref{tab: milu_stats} shows the overall statistics of \milu. Of the total 79K questions, 
only 25\% of questions are translated from English, with the remainder preserved in their original source languages. The average question length varies across languages, with languages such as \textit{kn}, \textit{ml}, and \textit{ta} having shorter word counts due to their agglutinative nature.

As shown in Figure \ref{fig:tag}, \milu~ consists of eight domains.

\noindent \textbf{Arts \& Humanities: }This domain covers topics such as Indian art, literature, dance, festivals, and architecture. Its content varies significantly by language, as it is highly dependent on regional culture.

\noindent \textbf{Science: } Includes topics in physics, chemistry, biology, with references to Ayurveda, and ancient as well as modern scientific achievements.

\noindent \textbf{Health \& Medicine: } Covers public health policies, modern healthcare, traditional Ayurveda and Unani medicine, and health-related government initiatives.

\noindent \textbf{Business Studies: }Focuses on entrepreneurship, trade, and economic policies such as Make in India. It also covers topics like taxation, MSMEs, and global trade.

\noindent \textbf{Law \& Governance: }Includes topics on the constitution, governance, judiciary, public administration, fundamental rights, and various governmental schemes, including local governance policies.

\noindent \textbf{Environmental Sciences: }Covers topics related to biodiversity, environmental policies, and both local and national environmental initiatives.

\noindent \textbf{Social Sciences: }Explores topics such as history, geography, and politics. The content in this domain is region-specific and varies by language.

\noindent \textbf{Engineering \& Technology: }Includes discussions on modern developments in India, such as IT, telecommunications, infrastructure, and space technology, with a focus on government policies. 

Figure \ref{fig:tag} shows the domain-wise statistics of \milu. We see a fairly large number of questions in the culturally relevant domains of Arts \& Humanities and Social Sciences. Detailed statistics per language can be found in the Appendix \ref{app: distrb}.

%% file: Sections/4_experiments.tex
\section{Experimental Setup}

We evaluate 42 different models on \milu, including large proprietary models, open-source multilingual models, and popular fine-tuned models specific to Indic languages. Both the base versions and instruction fine-tuned variants of these models, wherever applicable, are evaluated to measure the improvements gained from fine-tuning. All models, except for proprietary models and \textsc{Llama-3.1-405B}, are tested under 0-shot, 1-shot, and 5-shot setups. We maintain a separate validation set of approximately 9,000 questions to serve as examples for few-shot evaluations. 

For non-API-based models, we use the \textsc{LM-Evaluation-Harness}~\cite{eval-harness, Biderman2024LessonsFT} to ensure clean and reproducible evaluations. We use the log-likelihood method, where the probability of a given output string is computed by conditioning it on some provided input~\cite{gpt3}. Specifically, the log-likelihood of an answer $(a)$ given the question $(x)$, i.e., $log P(a|x)$, is calculated by concatenating the answer $(a)$ with question $(x)$, and then summing up the log probabilities, of each target token. For multiple choice questions, given $k$ possible answer strings, we select the answer string $(a_i)$ with the highest conditional log probability, i.e., $argmax(log P(a_1|x), ..., log P(a_k|x))$.

The API-based models are evaluated using the generative approach due to the lack of support for prompt log probabilities. We explicitly prompt these models to generate the correct response in a structured JSON format to simplify response parsing. Due to the high costs involved, these models are evaluated only in the zero-shot setup.

%% file: Sections/5_results.tex
\section{Results and Discussions}

\begingroup
\setlength{\tabcolsep}{6pt} 
\renewcommand{\arraystretch}{1} 
\begin{table*}[t]

\centering
\resizebox{\textwidth}{!}{%
\begin{tabular}{l|ccccccccccc|l}
\toprule
\textit{Model} & \textit{\textbf{bn}} & \textit{\textbf{en}} &  \textit{\textbf{gu}} &  \textit{\textbf{hi}} &  \textit{\textbf{kn}} &  \textit{\textbf{ml}} &  \textit{\textbf{mr}} &  \textit{\textbf{or}} &  \textit{\textbf{pa}} &  \textit{\textbf{ta}} &  \textit{\textbf{te}} &  \textit{\textbf{avg}} \\
\midrule
\multicolumn{13}{c}{\textbf{Closed Models}}  \\
\midrule
GPT-4o & \cellcolor[HTML]{FFD972}77.59 & \cellcolor[HTML]{FFD666}81.75 & \cellcolor[HTML]{FFDC79}74.64 & \cellcolor[HTML]{FFD96F}78.62 & \cellcolor[HTML]{FFDA73}76.93 & \cellcolor[HTML]{FFDD7F}72.57 & \cellcolor[HTML]{FFDC7D}73.44 & \cellcolor[HTML]{FFDF86}70.07 & \cellcolor[HTML]{FFDC7C}73.84 & \cellcolor[HTML]{FFDF85}70.15 & \cellcolor[HTML]{FFDD7F}72.53 & \cellcolor[HTML]{FFD666}74.74 \\
GPT-4o-mini & \cellcolor[HTML]{FFE69F}60.69 & \cellcolor[HTML]{FFDF84}70.52 & \cellcolor[HTML]{FFE8A6}57.84 & \cellcolor[HTML]{FFE498}63.14 & \cellcolor[HTML]{FFE8A9}56.92 & \cellcolor[HTML]{FFEAAF}54.52 & \cellcolor[HTML]{FFE9AC}55.76 & \cellcolor[HTML]{FFEFC2}47.31 & \cellcolor[HTML]{FFE8A6}57.89 & \cellcolor[HTML]{FFEDB9}50.84 & \cellcolor[HTML]{FFEBB2}53.31 & \cellcolor[HTML]{FFE59D}57.16 \\
Gemini-1.5-Pro & \cellcolor[HTML]{FFE089}68.8 & \cellcolor[HTML]{FFDA73}76.91 & \cellcolor[HTML]{FFDF87}69.68 & \cellcolor[HTML]{FFDE84}70.63 & \cellcolor[HTML]{FFDF87}69.68 & \cellcolor[HTML]{FFE292}65.5 & \cellcolor[HTML]{FFE292}65.63 & \cellcolor[HTML]{FFE49B}62.21 & \cellcolor[HTML]{FFE292}65.37 & \cellcolor[HTML]{FFE290}66.03 & \cellcolor[HTML]{FFE395}64.47 & \cellcolor[HTML]{FFDC7C}67.72 \\
Gemini-1.5-Flash & \cellcolor[HTML]{FFE59B}61.93 & \cellcolor[HTML]{FFDF87}69.42 & \cellcolor[HTML]{FFE59C}61.52 & \cellcolor[HTML]{FFE394}64.81 & \cellcolor[HTML]{FFE395}64.22 & \cellcolor[HTML]{FFE6A2}59.56 & \cellcolor[HTML]{FFE6A2}59.44 & \cellcolor[HTML]{FFE8A9}56.88 & \cellcolor[HTML]{FFE49B}62.23 & \cellcolor[HTML]{FFE7A3}58.89 & \cellcolor[HTML]{FFE7A6}58.13 & \cellcolor[HTML]{FFE18F}61.55 \\
Krutrim-spectre-v2 & \cellcolor[HTML]{FFF1CA}44.48 & \cellcolor[HTML]{FFF0C4}46.59 & \cellcolor[HTML]{FFF2CD}43.43 & \cellcolor[HTML]{FFF0C8}45.39 & \cellcolor[HTML]{FFF0C6}46.05 & \cellcolor[HTML]{FFF3D1}41.89 & \cellcolor[HTML]{FFF4D4}40.89 & \cellcolor[HTML]{FFF7DE}36.8 & \cellcolor[HTML]{FFF4D4}40.57 & \cellcolor[HTML]{FFF4D5}40.48 & \cellcolor[HTML]{FFF5D8}39.22 & \cellcolor[HTML]{FFF1CB}42.34 \\
\midrule
\multicolumn{13}{c}{\textbf{Open Multilingual Models}} \\
\midrule
\multicolumn{13}{c}{$<$ 4B Models} \\
\midrule
meta-llama/Llama-3.2-1B & \cellcolor[HTML]{FFFFFD}25.37 & \cellcolor[HTML]{FFF8E5}34.35 & \cellcolor[HTML]{FFFFFD}25.46 & \cellcolor[HTML]{FFFEF9}27.01 & \cellcolor[HTML]{FFFFFD}25.51 & \cellcolor[HTML]{FFFFFD}25.27 & \cellcolor[HTML]{FFFEFA}26.5 & \cellcolor[HTML]{FFFFFE}25.02 & \cellcolor[HTML]{FFFFFF}24.52 & \cellcolor[HTML]{FFFFFC}25.56 & \cellcolor[HTML]{FFFFFC}25.79 & \cellcolor[HTML]{FFFFFD}26.4 \\
meta-llama/Llama-3.2-1B-Instruct & \cellcolor[HTML]{FFFFFD}25.17 & \cellcolor[HTML]{FFFEFC}25.84 & \cellcolor[HTML]{FFFFFD}25.19 & \cellcolor[HTML]{FFFFFD}25.24 & \cellcolor[HTML]{FFFEFB}25.91 & \cellcolor[HTML]{FFFFFE}25.02 & \cellcolor[HTML]{FFFEFC}25.82 & \cellcolor[HTML]{FFFFFF}24.4 & \cellcolor[HTML]{FFFFFF}24.74 & \cellcolor[HTML]{FFFEF9}26.8 & \cellcolor[HTML]{FFFEFB}26.11 & \cellcolor[HTML]{FFFFFF}25.48 \\
sarvamai/sarvam-2b-v0.5 & \cellcolor[HTML]{FFFCF1}29.83 & \cellcolor[HTML]{FFF8E4}34.73 & \cellcolor[HTML]{FFFCF2}29.5 & \cellcolor[HTML]{FFFBED}31.37 & \cellcolor[HTML]{FFFCF3}29.03 & \cellcolor[HTML]{FFFEFB}25.94 & \cellcolor[HTML]{FFFCF3}28.96 & \cellcolor[HTML]{FFFEF8}27.09 & \cellcolor[HTML]{FFFCF3}29.06 & \cellcolor[HTML]{FFFDF7}27.68 & \cellcolor[HTML]{FFFDF4}28.57 & \cellcolor[HTML]{FFFCF4}29.25 \\
sarvamai/sarvam-1 & \cellcolor[HTML]{FFF9E8}33.31 & \cellcolor[HTML]{FFFBF0}30.3 & \cellcolor[HTML]{FFFCF1}29.83 & \cellcolor[HTML]{FFF9E8}33.05 & \cellcolor[HTML]{FFF9E8}33.29 & \cellcolor[HTML]{FFFBEE}30.8 & \cellcolor[HTML]{FFFDF6}27.9 & \cellcolor[HTML]{FFFCF2}29.33 & \cellcolor[HTML]{FFFCF4}28.76 & \cellcolor[HTML]{FFFDF7}27.64 & \cellcolor[HTML]{FFFDF5}28.5 & \cellcolor[HTML]{FFFCF1}30.25 \\
google/gemma-2-2b & \cellcolor[HTML]{FFFCF1}29.72 & \cellcolor[HTML]{FFEDB9}50.86 & \cellcolor[HTML]{FFFCF4}28.82 & \cellcolor[HTML]{FFFAEA}32.32 & \cellcolor[HTML]{FFFCF1}29.77 & \cellcolor[HTML]{FFFCF4}28.72 & \cellcolor[HTML]{FFFBEE}30.99 & \cellcolor[HTML]{FFFFFD}25.28 & \cellcolor[HTML]{FFFBEF}30.64 & \cellcolor[HTML]{FFFBF0}30.1 & \cellcolor[HTML]{FFFDF5}28.46 & \cellcolor[HTML]{FFFBED}31.43 \\
google/gemma-2-2b-it & \cellcolor[HTML]{FFFBEF}30.6 & \cellcolor[HTML]{FFEBB5}52.4 & \cellcolor[HTML]{FFFCF3}29.25 & \cellcolor[HTML]{FFF8E5}34.33 & \cellcolor[HTML]{FFFDF5}28.52 & \cellcolor[HTML]{FFFCF4}28.81 & \cellcolor[HTML]{FFFAEB}32.18 & \cellcolor[HTML]{FFFEFA}26.32 & \cellcolor[HTML]{FFFCF3}29.01 & \cellcolor[HTML]{FFFBEF}30.46 & \cellcolor[HTML]{FFFCF3}29.22 & \cellcolor[HTML]{FFFAEB}31.92 \\
meta-llama/Llama-3.2-3B & \cellcolor[HTML]{FFFAE9}32.75 & \cellcolor[HTML]{FFEDBB}50.25 & \cellcolor[HTML]{FFFBED}31.32 & \cellcolor[HTML]{FFF8E4}34.63 & \cellcolor[HTML]{FFFAEC}31.75 & \cellcolor[HTML]{FFFCF3}29.07 & \cellcolor[HTML]{FFFAEA}32.41 & \cellcolor[HTML]{FFFDF6}28.07 & \cellcolor[HTML]{FFFBEF}30.42 & \cellcolor[HTML]{FFFBEE}30.93 & \cellcolor[HTML]{FFFCF1}29.75 & \cellcolor[HTML]{FFF9E9}32.85 \\
meta-llama/Llama-3.2-3B-Instruct & \cellcolor[HTML]{FFF9E9}32.9 & \cellcolor[HTML]{FFFAEC}31.74 & \cellcolor[HTML]{FFF9E9}32.82 & \cellcolor[HTML]{FFFBEF}30.55 & \cellcolor[HTML]{FFFAEC}31.83 & \cellcolor[HTML]{FFFBF0}30.36 & \cellcolor[HTML]{FFFDF4}28.54 & \cellcolor[HTML]{FFFEFA}26.48 & \cellcolor[HTML]{FFFAEA}32.35 & \cellcolor[HTML]{FFFBED}31.32 & \cellcolor[HTML]{FFFBED}31.19 & \cellcolor[HTML]{FFFBEF}30.92 \\
nvidia/Nemotron-4-Mini-Hindi-4B-Base & \cellcolor[HTML]{FFFAED}31.41 & \cellcolor[HTML]{FFECB8}51.37 & \cellcolor[HTML]{FFFEFC}25.85 & \cellcolor[HTML]{FFECB7}51.42 & \cellcolor[HTML]{FFFDF7}27.4 & \cellcolor[HTML]{FFFEF9}27.01 & \cellcolor[HTML]{FFF5D8}39.37 & \cellcolor[HTML]{FFFEF9}26.9 & \cellcolor[HTML]{FFFEFA}26.47 & \cellcolor[HTML]{FFFDF6}27.89 & \cellcolor[HTML]{FFFDF5}28.25 & \cellcolor[HTML]{FFF9E8}33.03 \\

\midrule
\multicolumn{13}{c}{7B to 27B Models} \\
\midrule
Telugu-LLM-Labs/Navarasa-2.0 & \cellcolor[HTML]{FFF7DF}36.43 & \cellcolor[HTML]{FFECB7}51.69 & \cellcolor[HTML]{FFF6DD}37.44 & \cellcolor[HTML]{FFF4D6}39.86 & \cellcolor[HTML]{FFF4D6}40.01 & \cellcolor[HTML]{FFF7DF}36.40 & \cellcolor[HTML]{FFF6DE}37.00 & \cellcolor[HTML]{FFF8E2}35.36 & \cellcolor[HTML]{FFF7DE}36.86 & \cellcolor[HTML]{FFF6DD}37.18 & \cellcolor[HTML]{FFF7DE}36.86 & \cellcolor[HTML]{FFF5D7}38.64 \\
neulab/Pangea-7B & \cellcolor[HTML]{FFF3D0}42.06 & \cellcolor[HTML]{FFE9AA}56.41 & \cellcolor[HTML]{FFF7DE}36.81 & \cellcolor[HTML]{FFF3D0}42.21 & \cellcolor[HTML]{FFF9E6}33.83 & \cellcolor[HTML]{FFF9E6}33.86 & \cellcolor[HTML]{FFF7DF}36.73 & \cellcolor[HTML]{FFFAEB}32.18 & \cellcolor[HTML]{FFF7DF}36.45 & \cellcolor[HTML]{FFFAE9}32.72 & \cellcolor[HTML]{FFF7DF}36.43 & \cellcolor[HTML]{FFF5D8}38.15 \\
CohereForAI/aya-23-8B & \cellcolor[HTML]{FFFDF6}27.85 & \cellcolor[HTML]{FFF0C7}45.76 & \cellcolor[HTML]{FFFEF9}26.93 & \cellcolor[HTML]{FFF7E0}36.31 & \cellcolor[HTML]{FFFDF7}27.53 & \cellcolor[HTML]{FFFDF5}28.4 & \cellcolor[HTML]{FFFBEE}30.96 & \cellcolor[HTML]{FFFDF7}27.51 & \cellcolor[HTML]{FFFEF9}26.98 & \cellcolor[HTML]{FFFCF4}28.88 & \cellcolor[HTML]{FFFEFA}26.34 & \cellcolor[HTML]{FFFBF0}30.31 \\
meta-llama/Llama-3.1-8B & \cellcolor[HTML]{FFF6DC}37.89 & \cellcolor[HTML]{FFE6A1}59.66 & \cellcolor[HTML]{FFF8E3}35.26 & \cellcolor[HTML]{FFF2CF}42.69 & \cellcolor[HTML]{FFF6DE}37.09 & \cellcolor[HTML]{FFF8E4}34.71 & \cellcolor[HTML]{FFF6DC}37.65 & \cellcolor[HTML]{FFF9E9}32.86 & \cellcolor[HTML]{FFF7E0}36.33 & \cellcolor[HTML]{FFF7E0}36.25 & \cellcolor[HTML]{FFF8E3}34.94 & \cellcolor[HTML]{FFF5D7}38.67 \\
meta-llama/Llama-3.1-8B-Instruct & \cellcolor[HTML]{FFF8E4}34.53 & \cellcolor[HTML]{FFF0C5}46.22 & \cellcolor[HTML]{FFF9E7}33.77 & \cellcolor[HTML]{FFF7E1}35.73 & \cellcolor[HTML]{FFF7E0}36.04 & \cellcolor[HTML]{FFF8E5}34.37 & \cellcolor[HTML]{FFF9E8}33.28 & \cellcolor[HTML]{FFFBF0}30.08 & \cellcolor[HTML]{FFF9E8}33.33 & \cellcolor[HTML]{FFF9E5}34.15 & \cellcolor[HTML]{FFF9E7}33.6 & \cellcolor[HTML]{FFF8E2}35.01 \\
google/gemma-2-9b & \cellcolor[HTML]{FFEBB3}53.12 & \cellcolor[HTML]{FFDF86}69.8 & \cellcolor[HTML]{FFECB8}51.19 & \cellcolor[HTML]{FFE9AB}56.24 & \cellcolor[HTML]{FFEAB1}53.99 & \cellcolor[HTML]{FFEEBC}49.55 & \cellcolor[HTML]{FFEDBC}49.81 & \cellcolor[HTML]{FFF4D5}40.42 & \cellcolor[HTML]{FFEDBC}49.7 & \cellcolor[HTML]{FFEDBB}50.05 & \cellcolor[HTML]{FFEEBE}48.78 & \cellcolor[HTML]{FFE9AD}52.06 \\
google/gemma-2-9b-it & \cellcolor[HTML]{FFF0C6}46.04 & \cellcolor[HTML]{FFE397}63.61 & \cellcolor[HTML]{FFF5D9}38.82 & \cellcolor[HTML]{FFF1C9}44.82 & \cellcolor[HTML]{FFF2CD}43.47 & \cellcolor[HTML]{FFF4D4}40.8 & \cellcolor[HTML]{FFF6DC}37.62 & \cellcolor[HTML]{FFF9E5}34.19 & \cellcolor[HTML]{FFF4D5}40.42 & \cellcolor[HTML]{FFF5D9}38.83 & \cellcolor[HTML]{FFF5D7}39.44 & \cellcolor[HTML]{FFF1CA}42.55 \\
google/gemma-2-27b & \cellcolor[HTML]{FFE497}63.47 & \cellcolor[HTML]{FFDB79}74.82 & \cellcolor[HTML]{FFE7A2}59.29 & \cellcolor[HTML]{FFE395}64.33 & \cellcolor[HTML]{FFE497}63.43 & \cellcolor[HTML]{FFE6A2}59.62 & \cellcolor[HTML]{FFE6A1}59.66 & \cellcolor[HTML]{FFF1C9}44.86 & \cellcolor[HTML]{FFE6A1}59.94 & \cellcolor[HTML]{FFE7A4}58.6 & \cellcolor[HTML]{FFE8A6}57.94 & \cellcolor[HTML]{FFE293}60.54 \\
google/gemma-2-27b-it & \cellcolor[HTML]{FFE6A1}59.82 & \cellcolor[HTML]{FFDC7C}73.7 & \cellcolor[HTML]{FFEAAD}55.17 & \cellcolor[HTML]{FFE59D}61.25 & \cellcolor[HTML]{FFE7A5}58.23 & \cellcolor[HTML]{FFEBB2}53.37 & \cellcolor[HTML]{FFEAAE}54.78 & \cellcolor[HTML]{FFF1C9}44.99 & \cellcolor[HTML]{FFE9AC}55.55 & \cellcolor[HTML]{FFE9AC}55.6 & \cellcolor[HTML]{FFEAB0}54.29 & \cellcolor[HTML]{FFE59E}56.98 \\

\midrule
\multicolumn{13}{c}{$>$ 27B Models} \\
\midrule
CohereForAI/aya-23-35B & \cellcolor[HTML]{FFF9E9}32.9 & \cellcolor[HTML]{FFEBB3}53.09 & \cellcolor[HTML]{FFFBF0}30.35 & \cellcolor[HTML]{FFF2CC}43.53 & \cellcolor[HTML]{FFFBED}31.39 & \cellcolor[HTML]{FFFAEA}32.42 & \cellcolor[HTML]{FFF6DD}37.22 & \cellcolor[HTML]{FFFBED}31.27 & \cellcolor[HTML]{FFFDF8}27.27 & \cellcolor[HTML]{FFF8E3}35 & \cellcolor[HTML]{FFFCF2}29.34 & \cellcolor[HTML]{FFF8E2}34.89 \\
meta-llama/Llama-3.1-70B & \cellcolor[HTML]{FFE18D}67.37 & \cellcolor[HTML]{FFDC7A}74.59 & \cellcolor[HTML]{FFE49A}62.42 & \cellcolor[HTML]{FFE18E}66.97 & \cellcolor[HTML]{FFE08A}68.37 & \cellcolor[HTML]{FFE499}62.69 & \cellcolor[HTML]{FFE49B}62.2 & \cellcolor[HTML]{FFE8A7}57.48 & \cellcolor[HTML]{FFE394}64.6 & \cellcolor[HTML]{FFE69F}60.7 & \cellcolor[HTML]{FFE69F}60.64 & \cellcolor[HTML]{FFDF87}64.37 \\
meta-llama/Llama-3.1-70B-Instruct & \cellcolor[HTML]{FFE499}62.77 & \cellcolor[HTML]{FFEAB1}54.00 & \cellcolor[HTML]{FFE6A0}60.26 & \cellcolor[HTML]{FFE7A3}58.90 & \cellcolor[HTML]{FFE498}63.02 & \cellcolor[HTML]{FFE7A3}58.94 & \cellcolor[HTML]{FFEBB1}53.71 & \cellcolor[HTML]{FFEBB4}52.81 & \cellcolor[HTML]{FFE69F}60.67 & \cellcolor[HTML]{FFE7A5}58.41 & \cellcolor[HTML]{FFE8A7}57.66 & \cellcolor[HTML]{FFE49A}58.29 \\
meta-llama/Llama-3.1-405B & \cellcolor[HTML]{FFE293}65.15 & \cellcolor[HTML]{FFE59D}61.42 & \cellcolor[HTML]{FFE59E}61.11 & \cellcolor[HTML]{FFECB9}50.98 & \cellcolor[HTML]{FFE397}63.69 & \cellcolor[HTML]{FFE397}63.64 & \cellcolor[HTML]{FFEDBB}50.21 & \cellcolor[HTML]{FFE9AD}55.35 & \cellcolor[HTML]{FFE59C}61.84 & \cellcolor[HTML]{FFE7A4}58.81 & \cellcolor[HTML]{FFE6A1}59.67 & \cellcolor[HTML]{FFE397}59.26 \\
\bottomrule
\end{tabular}%
}
\caption{Evaluation results of all the  Closed and Open Multilingual models supporting all languages on \milu. We report 5-shot accuracies for all open models (except for \textsc{Llama-3.1-70B-Instruct} \& \textsc{Llama-3.1-405B} for which we report 0-shot accuracy) with the accuracy averaged across all the domains per language. Higher values indicate better model performance for the given language. Refer Section (\S\ref{sec: overal}) for more details.}
\label{tab:main}
\end{table*}
\endgroup

In this section, we discuss the results and our findings across all the experiments conducted.

\subsection{How do models supporting all languages perform?}
\label{sec: overal}

Table \ref{tab:main} shows the performance of various models supporting all the languages in the 5-shot setup. Among these, \textsc{GPT-4o} emerges as the strongest model, consistently outperforming all competitors across languages, with an average accuracy of 74.74\%. Closed proprietary models, generally achieve higher performance than the open-source alternatives. Among the open-source models, \textsc{Llama-3.1-70B} shows the best performance. 

Performance is notably higher in high-resource languages- \textit{en}, \textit{hi}, \textit{bn} -as compared to the low-resource ones. Interestingly, the models also show low accuracies in agglutinative languages, such as \textit{ta}, \textit{te}. Notably, models specifically trained for Indian languages such as \textsc{Krutrim-spectre-v2} and \textsc{sarvam-2b-v0.5}, perform slightly below their English counterparts of comparable scale. Additionally, while several models achieve better results in English, their performance in other languages is often nearly at the random baseline.

\subsection{How do language-specific fine-tuned models perform?}
\label{sec: lang_specific}

\begingroup
\setlength{\tabcolsep}{3pt} 
\renewcommand{\arraystretch}{1} 
\begin{table*}[t]

\centering
\resizebox{\textwidth}{!}{%
\begin{tabular}{l|l|ccccccccc}
\toprule
\textit{\textbf{lang}} & \textbf{Model} & 
\textbf{\begin{tabular}[c]{@{}c@{}}Business \\ Studies\end{tabular}} & \textbf{\begin{tabular}[c]{@{}c@{}}Engg. \\ \& Tech\end{tabular}} & \textbf{\begin{tabular}[c]{@{}c@{}}Social \\ Sciences\end{tabular}} & \textbf{\begin{tabular}[c]{@{}c@{}}Env. \\ Sciences\end{tabular}} & \textbf{\begin{tabular}[c]{@{}c@{}}Law \&\\  Governance\end{tabular}} & \textbf{\begin{tabular}[c]{@{}c@{}}Science\end{tabular}} & \textbf{\begin{tabular}[c]{@{}c@{}}Arts \& \\  Humanities\end{tabular}} & \textbf{\begin{tabular}[c]{@{}c@{}}Health \& \\ Medicine\end{tabular}} & \multicolumn{1}{l}{\textbf{Avg}} \\
\midrule
\multirow{6}{*}{\textit{\textbf{hi}}}   & OpenHathi-7B-Hi   & 29.72 & 28.59 & 28.81 & 27.74 & 30.96 & 28.53 & 28.28 & 30.76 & 29.17 \\
 & Airavata  & 28.03 & 28.45 & 28.92 & 27.14 & 27.59 & 28.95 & 27.54 & 32.77 & 28.67 \\
 & ProjectIndus  & 26.21 & 25.83 & 26.95 & 25.03 & 22.93 & 24.91 & 25.13 & 24.41 & 25.18 \\
 & Gajendra-v0.1 & 29.30 & 28.36 & 25.54 & 29.30 & 27.75 & 29.18 & 28.38 & 30.26 & 28.51 \\
 & open-aditi-v6-llama3  & 29.30 & 31.65 & 29.99 & 30.15 & 34.78 & 32.57 & 32.43 & 37.12 & 32.25 \\
 & AryaBhatta-GemmaGenZ-Merged & \textbf{39.42} & \textbf{37.88} & \textbf{37.59} & \textbf{41.14} & \textbf{39.60} & \textbf{41.25} & \textbf{35.71} & \textbf{48.49} & \textbf{40.14} \\
 \midrule
\multirow{3}{*}{\textit{\textbf{te}}}  & telugu-llama-7b-instruct-v0.1 & \textbf{30.19} & \textbf{27.20} & \textbf{26.27} & 24.03 & 27.16 & 26.41 & 26.81 & 24.63 & \textbf{26.59} \\
 & Telugu-Llama2-7B-v0-Base  & 22.96 & 27.10 & 24.80 & \textbf{27.01} & \textbf{26.14} & \textbf{27.13} & 25.42 & 27.53 & 26.01 \\
 & Telugu-Llama2-7B-v0-Instruct  & 26.03 & 24.00 & 24.45 & 23.68 & 24.48 & 26.82 & \textbf{27.17} & \textbf{28.01} & 25.58 \\
 \midrule
\multirow{3}{*}{\textit{\textbf{ka}}} & Kan-LLaMA-7B-base & 29.00 & 27.42 & \textbf{30.37} & 27.42 & \textbf{28.47} & 26.92 & \textbf{29.28} & 30.50 & 28.67 \\
 & Ambari-7B-base-v0.1   & \textbf{31.49} & 27.52 & 29.08 & \textbf{28.70} & 27.15 & \textbf{27.82} & 27.93 & \textbf{31.00} & \textbf{28.84} \\
 & Ambari-7B-Instruct-v0.1   & 28.17 & \textbf{28.02} & 24.83 & 27.78 & 23.84 & 27.56 & 27.42 & 27.00 & 26.83 \\
 \midrule
\multirow{2}{*}{\textit{\textbf{mr}}} & Misal-7B-base-v0.1 & \textbf{26.96} & \textbf{26.65} & \textbf{25.02} & \textbf{27.19} & \textbf{27.28} & \textbf{26.11} & \textbf{26.82} & \textbf{24.05} & \textbf{26.26} \\
 & Misal-7B-instruct-v0.1 & 25.05 & 21.44 & 21.55 & 22.89 & 21.82 & 24.59 & 23.36 & 22.16 & 22.86 \\
 \midrule
\textit{\textbf{ta}} & tamil-llama-7b-instruct-v0.2  & 22.22 & 24.12 & 25.32 & 24.89 & 28.69 & 24.98 & 25.07 & 22.79 & 24.76 \\
\midrule
\textit{\textbf{ml}} & malayalam-llama-7b-instruct-v0.1  & 30.00 & 27.92 & 23.42 & 27.88 & 25.00 & 27.57 & 23.33 & 20.00 & 25.64 \\
\bottomrule
\end{tabular}%
}
\caption{Evaluation results of all the language-specific fine-tuned models on \milu. We report domain level 5-shot accuracies for all the models on the language supported by the model. Higher values indicate better model performance for the given domain. Refer to Section (\S\ref{sec: lang_specific}) for more details.}
\label{tab:langspecific}
\end{table*}
\endgroup

We evaluate around 16 Indic language LLMs on \milu. These models are primarily built by adapting English LLMs, such as \textsc{Llama-2-7B}, by first continually pretraining on small amount Indic language data, followed by optionally instruction fine-tuning them. As seen from Table \ref{tab:langspecific}, across languages, the models exhibit average performance comparable to random baselines, with minimal variations among them.

Among the evaluated models, the \textsc{AryaBhatta-GemmaGenZ-Vikas} model stands out, performing better than its counterparts. When analyzed across domains, the models generally perform worse in Arts, Humanities, and Social Sciences than in STEM subjects.
\begin{figure}[h]
    \centering
    \includegraphics[width=\columnwidth]{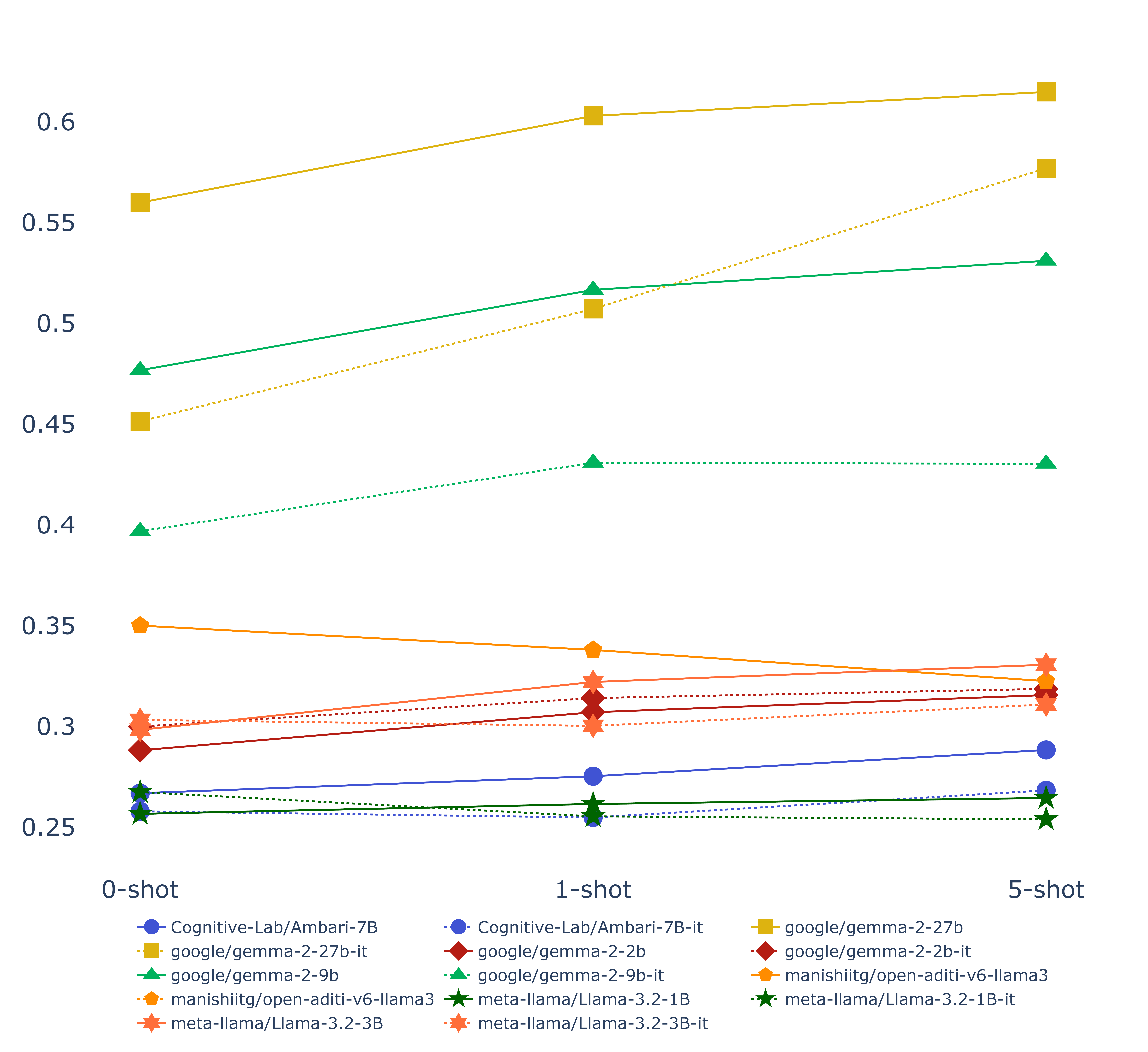}
    \caption{Comparison of Base and Instruct models averaged across all languages for varying number of in-context examples. We plot the average accuracies of the \textsc{Gemma} and \textsc{Llama} series of models, highlighting the performance trend as the number of in-context examples increases. Refer to Section (\S\ref{sec: shots}) for more details.}
    \label{fig:shot}
\end{figure}

\begin{figure*}[ht!]
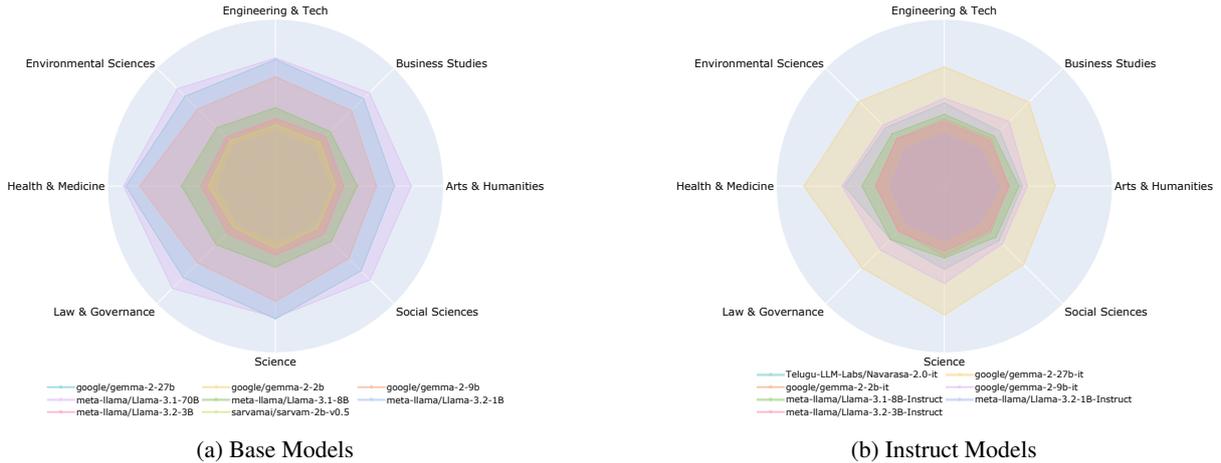

    \centering
    \begin{subcaptionbox}{Base Models\label{fig:img1}}{%
   \includegraphics[width=0.45\textwidth]{Figures/base_models_performance.pdf}%
    }\end{subcaptionbox}
    \hfill
    \begin{subcaptionbox}{Instruct Models\label{fig:img2}}{%
   \includegraphics[width=0.45\textwidth]{Figures/instruction-tuned_models_performance.pdf}%
    }\end{subcaptionbox}
    \caption{Evaluation results of Base models (\ref{fig:img1}) and Instruct models (\ref{fig:img2}) on the different domains supported in \milu. The plot shows the average 5-shot accuracies across all languages for various models. Refer to Section (\S\ref{sec: domains})}
    \label{fig:side-by-side}
\end{figure*}
\subsection{How do the models respond to In-Context examples?}
\label{sec: shots}

We compare the performance of different Base and Instruct models across zero, one, and five-shot setups. As shown in Figure \ref{fig:shot}, the performance of base models consistently improves with an increasing number of in-context examples, with the 5-shot setup yielding the best results. In contrast, Instruct models exhibit more varied behavior, where models either stagnate or even degrade in performance. This also aligns with expectations, as Instruct models are specifically fine-tuned as conversation assistants and may not respond well to the few-shot in-context examples format. This also correlates with results in Table~\ref{tab:main} where most instruction finetuned variants performs worse on English itself. This essentially implies the inferior performance is less about the language shift but due to the change of tasks during instruction finetuning and inference task for \milu evaluation. Understandably, the instruction finetuned variants performs poor on the overall average across languages as well.

\subsection{How does the performance vary with scale?}
\label{sec: scale}
\begin{figure}[h]
    \centering
    \includegraphics[width=\columnwidth]{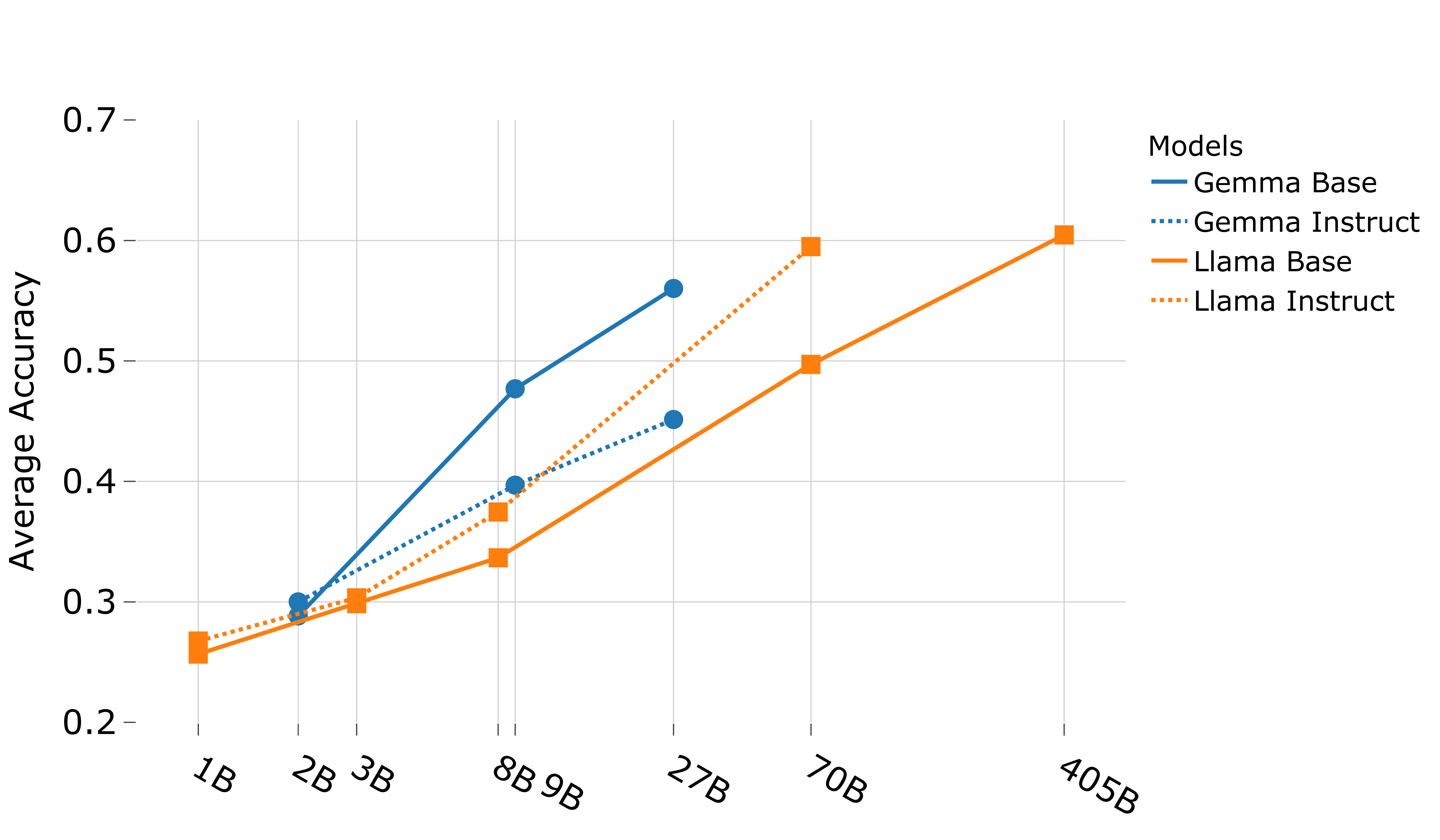}
    \caption{Comparison of performance of \textsc{Gemma} and \textsc{Llama} family of models across different parameter scales. We plot the zero-shot average accuracies of all models across languages. Refer to Section (\S\ref{sec: scale}) for more details.}
    \label{fig:scale}
\end{figure}

We evaluate the \textsc{Llama} and \textsc{Gemma} family of models, ranging from 1B to 405B parameters, to analyze how performance scales with model size. Figure \ref{fig:scale} shows that the model performance improves significantly with increasing scale. Notably, instruction-tuned models in the \textsc{Llama} family show more substantial improvements as compared to those in the \textsc{Gemma} family.

\subsection{How do the models perform on different domains?}
\label{sec: domains}

We analyze the performance of various base and instruct models across multiple domains and languages. Similar trends to those in Section (\S\ref{sec: lang_specific}) are observed where the open models perform poorly in domains specific to Indian culture—such as Arts \& Humanities, Social Sciences, and Law \& Governance—but demonstrate higher performance in STEM fields. This suggests that the training corpora for these models lack sufficient culturally specific data. Bridging this gap requires a more inclusive data distribution that ensures equitable representation of all cultures and languages.

\begingroup
\setlength{\tabcolsep}{3pt} 
\renewcommand{\arraystretch}{1} 
\begin{table}[t!]

\scriptsize
\centering
\begin{tabular}{llllccc}
\toprule
\textit{\textbf{lang}} & \textbf{Model}& \textbf{CPT} & \textbf{IFT} & \textbf{0}  & \textbf{1}  & \textbf{5}  \\
\midrule
\multirow{3}{*}{\textit{hi}} & OpenHathi & \textcolor{pert_green}{Y}   & \textcolor{pert_red}{N}   & -0.28\% & -0.25\% & 0.58\% \\
& Airavata  & \textcolor{pert_green}{Y}   & \textcolor{pert_green}{Y}   & \textbf{1.70\%} & \textbf{2.40\%} & 3.11\% \\
& Gajendra-v0.1 & \textcolor{pert_green}{Y}   & \textcolor{pert_green}{Y}   & 0.73\% & 0.92\% & \textbf{3.22\%} \\
\midrule
\multirow{3}{*}{\textit{kn}} & Ambari-base   & \textcolor{pert_green}{Y}   & \textcolor{pert_red}{N}   & \textbf{0.72\%} & 1.17\% & 2.46\% \\
& Ambari-Instruct& \textcolor{pert_green}{Y}   & \textcolor{pert_green}{Y}   & 0.60\% & 0.49\% & 1.87\% \\
& Kan-7B-base   & \textcolor{pert_green}{Y}   & \textcolor{pert_red}{N}   & \textbf{1.13\%} & \textbf{0.84\%} & \textbf{1.30\%} \\
\midrule
\multirow{3}{*}{\textit{te}} & telugu-llama-instruct    & \textcolor{pert_green}{Y}   & \textcolor{pert_green}{Y}   & \textbf{1.13\%} & \textbf{0.84\%} & \textbf{1.30\%} \\  
& Telugu-7B-v0-Base   & \textcolor{pert_green}{Y}   & \textcolor{pert_red}{N}   & 0.32\% & -0.52\% & 0.88\% \\
& Telugu-7B-v0-Instruct    & \textcolor{pert_green}{Y}   & \textcolor{pert_green}{Y}   & -0.98\% & -0.12\% & 0.29\% \\
\midrule
\multirow{2}{*}{\textit{mr}} & Misal-7B-base-v0.1  & \textcolor{pert_green}{Y}   & \textcolor{pert_red}{N}   & \textbf{0.16\%} & \textbf{-0.60\%} & \textbf{-0.93\%} \\
& Misal-7B-instruct-v0.1   & \textcolor{pert_green}{Y}   & \textcolor{pert_green}{Y}   & -2.55\% & -2.17\% & -2.38\% \\
\midrule
\textit{ta}   & tamil-llama-instruct& \textcolor{pert_green}{Y}   & \textcolor{pert_green}{Y}   & -0.06\% & -0.04\% & -0.19\% \\
\midrule
\textit{ml}   & malayalam-llama-instruct & \textcolor{pert_green}{Y}   & \textcolor{pert_green}{Y}   & 0.77\% & 1.09\% & 0.33\% \\
\bottomrule
\end{tabular}
\caption{Performance differences of Indian language models compared to the \textsc{Llama-2-7B} baseline at different in-context example settings. Each value shows the relative improvement or decline (\% difference) over the baseline, with a +ve value indicating improvement and a -ve value indicating a drop in the performance. Refer to Section (\S\ref{sec: lang_adap}) for more details.}
\label{tab: llama_comp}
\end{table}

\subsection{Does language adaptation on models help?}
\label{sec: lang_adap}
As most Indic LLMs are built on English base models like \textsc{Llama-2-7b}, we assess the impact of language adaptation on their performance. Table \ref{tab: llama_comp} compares language-specific models with the original \textsc{Llama-2-7b}, and instruction-tuned models with \textsc{Llama-2-7b-chat}. Our findings show minimal gains, with some models even underperforming post-adaptation.

Given the varied training data, this is not a direct "apples-to-apples" comparison but highlights key challenges. We conjecture that limited performance gains may result from small language-specific datasets and reliance on parameter-efficient methods like LoRA \cite{lora}. Another contributing factor could be the lack of diversity in instruction fine-tuning datasets. Models like \textsc{Airavata}, which utilize more diverse data \cite{gala2024airavata}, exhibit noticeably better performance. Further investigation is required to fully understand the limitations and opportunities in this area.


%% file: Sections/conclusion.tex
\section{Conclusion}
In this paper, we introduced \milu{}—\textbf{M}ultilingual \textbf{I}ndic \textbf{L}anguage \textbf{U}nderstanding Benchmark-a comprehensive benchmark specifically designed to evaluate LLMs across 11 Indic languages, spanning diverse domains and culturally relevant subjects. We evaluate 45 different LLMs and find that the majority of LLMs struggle on \milu, with GPT-4o achieving the highest average accuracy. The analysis also shows that models perform significantly better in high-resource languages than low-resource ones, highlighting the need for more robust multilingual strategies. Additionally, the domain-specific analysis indicates that models perform better in general fields such as STEM while facing challenges in culturally relevant subjects like Arts, Humanities, and Law, highlighting the lack of this knowledge in the current models and datasets. To foster open research, we release \milu{} alongside all code, and other artifacts. As LLMs continue to become pivotal in modern applications, we hope that \milu{} offers a foundational benchmark for developing more inclusive, culturally aware models that perform well across both general and culturally relevant domains.

%% file: Sections/limits.tex
\section*{Limitations}
This work has a few limitations. First, we restricted our study to the top 11 languages due to the lack of readily available questions in low-resource languages, which we aim to address in future work. Second, limited computational resources prevented a thorough evaluation of larger models, such as \textsc{Llama-3.1-70B-Instruct} and \textsc{Llama-3.1-405B}. Third, the scarcity of questions necessitated translating a portion of the dataset. Finally, our evaluation primarily relies on the log-likelihood approach, which may yield different results compared to other established methods, such as generation-based evaluation and chain-of-thought (CoT) prompting.

%% file: Sections/acknowledgments.tex
\section*{Acknowledgements}
We would like to thank EkStep Foundation and Nilekani Philanthropies for their generous grant towards building datasets, models, tools and other resources for Indian languages. We are also immensely grateful to the volunteers from the AI4Bharat team for their motivation and meticulous efforts in conducting manual audits.

%% file: Sections/ethics.tex
\section*{Ethics}
All data described in this work was scraped from publicly available resources. The datasets used in this paper will be made available under permissible licenses. Additionally, the code used for our evaluations will be made publicly available under the MIT License. We only used ChatGPT for assistance purely with the language of the paper, e.g., paraphrasing, spell-checking, or polishing the author’s original content, without suggesting new content.

%% file: Sections/appendix.tex
\section*{Appendix}

\section{Details about different Exams}
\label{app: exams}
We collected our questions from over 40 exam types ranging from various National and state level civil service examinations to examinations conducted by various government and private organizations. Tables \ref{tab:national_exams}, \ref{tab:orgs} and \ref{tab:civil_service_exams}

\begin{table*}[]
 \centering
 \begin{tabular}{l|c}
  \hline
  \textbf{Organization} & \textbf{Years Covered} \\
  \hline
  Railway Recruitment Board & 2018-2019 \\
  Railway Recruitment Cell & 2018 \\
  Dedicated Freight Corridor Corporation of India Limited & 2016-2021 \\
  Intelligence Bureau & 2012-2021 \\
  Union Public Service Commission & 2010-2023 \\
  \hline
 \end{tabular}
 \caption{Overview of various national-level exams and the corresponding years of coverage considered in \milu.}
 \label{tab:national_exams}
\end{table*}

\begin{table*}[]
 \centering
 \begin{tabular}{l|c}
  \hline
  \textbf{Organization} & \textbf{Years Covered} \\
  \hline
  Punjab Police & 2016-2022 \\
  Punjab State Power Corporation Limited & 2018 \\
  Chandigarh Police & 2018 \\
  Telangana Police & 2015-2022 \\
  Andhra Pradesh Police & 2016-2018 \\
  Tamil Nadu Public Service Commission & 2010-2021 \\
  Tamil Nadu Uniformed Services Recruitment Board & 2010-2022 \\
  Odisha Police & 2022 \\
  Karnataka State Police & 2019-2023 \\
  Madhya Pradesh Police & 2016-2023 \\
  Delhi Police & 2014-2022 \\
  Haryana Police & 2021 \\
  Lucknow Metro Rail Corporation & 2018 \\
  Delhi Metro Rail Corporation & 2017-2020 \\
  Punjab Subordinate Service Selection Board & 2022 \\
  Punjab State Transmission Corporation Limited & 2016 \\
  Gujarat Metro Rail Corporation & 2022 \\
  Kolkata Police & 2023 \\
  Maharashtra Police & 2021 \\
  Jharkhand Police & 2015-2021 \\
 \hline
 \end{tabular}
 \caption{Overview of various government and private organization exams and the corresponding years of coverage considered in \milu.}
 \label{tab:orgs}
\end{table*}

\begin{table*}[]
 \centering
 \begin{tabular}{l|c}
  \hline
  \textbf{Organization} & \textbf{Years Covered} \\
  \hline
  Staff Selection Commission & 2017-2023 \\
  West Bengal Civil Service & 2015-2021 \\
  Bihar Public Service Commission & 2015-2020 \\
  Maharashtra Public Service Commission & 2013-2023 \\
  Rajasthan Subordinate and Ministerial Services Selection Board & 2016-2022 \\
  Odisha Public Service Commission & 2016-2022 \\
  Uttar Pradesh Public Service Commission & 2012-2023 \\
  Haryana Public Service Commission & 2014-2022 \\
  Andhra Pradesh Public Service Commission & 2017-2019 \\
  Chhattisgarh Public Service Commission & 2014-2022 \\
  Jammu \& Kashmir Public Service Commission & 2021 \\
  Himachal Pradesh Public Service Commission & 2015-2022 \\
  Jharkhand Public Service Commission & 2021 \\
  Delhi Subordinate Services Selection Board & 2019-2022 \\
  \hline
 \end{tabular}
 \caption{Overview of various State-level civil services exams and the corresponding years of coverage considered in \milu.}
 \label{tab:civil_service_exams}
\end{table*}

\section{Details about subject and language distribution.}
\label{app: distrb}

\begin{figure*}[]
 \centering
 \includegraphics[width=\textwidth]{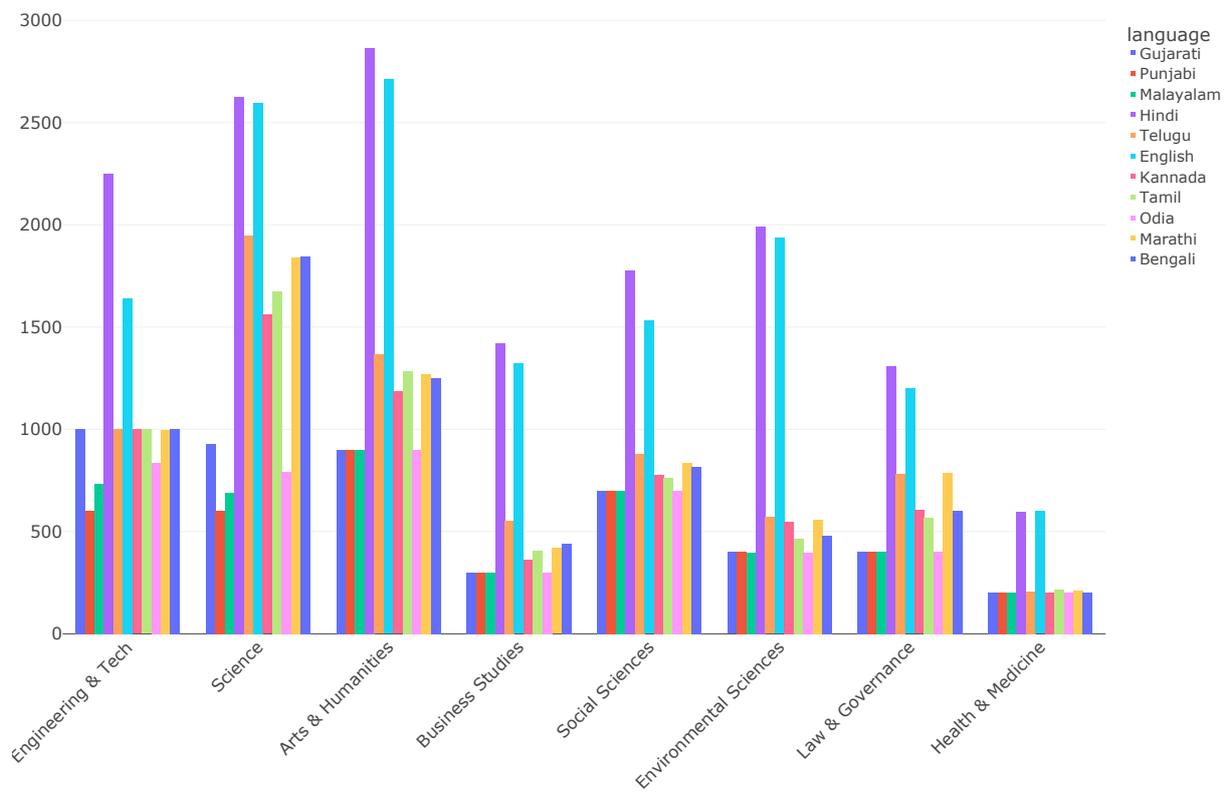}
 \caption{Distribution of tags across languages in MILU.}
 \label{fig: lang_disrib}
\end{figure*}

Detailed analysis of topic and language distribution across languages can be found in Table \ref{tab:detailed_stats} and Figure \ref{fig: lang_disrib}

\begingroup
\setlength{\tabcolsep}{3pt} 
\renewcommand{\arraystretch}{1} 
\begin{table*}[t]

\centering
\resizebox{\textwidth}{!}{%
\begin{tabular}{l|ccccccccccc|c}
\toprule
\textbf{Topic} & \multicolumn{1}{c}{\textit{\textbf{bn}}} & \multicolumn{1}{c}{\textit{\textbf{en}}} & \multicolumn{1}{c}{\textit{\textbf{gu}}} & \multicolumn{1}{c}{\textit{\textbf{hi}}} & \multicolumn{1}{c}{\textbf{\textit{kn}}} & \multicolumn{1}{c}{\textbf{\textit{ml}}} & \multicolumn{1}{c}{\textbf{\textit{mr}}} & \multicolumn{1}{c}{\textbf{\textit{or}}} & \multicolumn{1}{c}{\textbf{\textit{pa}}} & \multicolumn{1}{c}{\textbf{\textit{ta}}} & \multicolumn{1}{c}{\textbf{\textit{te}}} & \multicolumn{1}{c}{\textbf{Total}}\\
\midrule
Agriculture & 100 & 497 & 99 & 497 & 100 & 99 & 100 & 100 & 100 & 96 & 99 & 1887 \\
Anthropology & 100 & 100 & 100 & 99 & 100 & 100 & 100 & 100 & 100 & 98 & 100 & 1097 \\
Architecture and Design & 100 & 100 & 100 & 100 & 100 & 100 & 100 & 100 & 100 & 100 & 100 & 1100 \\
Arts and Culture & 165 & 500 & 100 & 499 & 163 & 100 & 183 & 100 & 100 & 157 & 199 & 2266 \\
Astronomy and Astrophysics & 100 & 100 & 100 & 128 & 100 & 100 & 100 & 100 & 100 & 100 & 100 & 1128 \\
Biology & 284 & 499 & 100 & 499 & 244 & 99 & 335 & 100 & 100 & 314 & 318 & 2892 \\
Business and Management & 100 & 323 & 100 & 427 & 100 & 100 & 100 & 100 & 100 & 100 & 106 & 1656 \\
Chemistry & 364 & 499 & 116 & 499 & 306 & 100 & 379 & 104 & 100 & 337 & 401 & 3205 \\
Computer Science & 192 & 500 & 100 & 500 & 169 & 100 & 169 & 100 & 100 & 172 & 223 & 2325 \\
Defense and Security & 100 & 100 & 100 & 211 & 100 & 100 & 100 & 100 & 100 & 100 & 100 & 1211 \\
Earth Sciences & 100 & 440 & 100 & 499 & 100 & 100 & 100 & 99 & 100 & 100 & 100 & 1838 \\
Economics & 139 & 498 & 100 & 496 & 100 & 100 & 142 & 100 & 100 & 126 & 167 & 2068 \\
Education & 100 & 314 & 100 & 338 & 100 & 100 & 157 & 100 & 100 & 99 & 100 & 1608 \\
Energy and Power & 100 & 148 & 100 & 295 & 100 & 100 & 100 & 100 & 100 & 100 & 100 & 1343 \\
Engineering & 500 & 500 & 500 & 500 & 500 & 234 & 500 & 336 & 100 & 499 & 500 & 4669 \\
Environmental Science & 100 & 499 & 100 & 499 & 128 & 100 & 190 & 99 & 100 & 99 & 179 & 2093 \\
Ethics and Human Rights & 100 & 100 & 100 & 100 & 100 & 100 & 100 & 100 & 100 & 100 & 100 & 1100 \\
Finance and Investment & 202 & 500 & 100 & 500 & 162 & 100 & 177 & 100 & 100 & 179 & 280 & 2400 \\
Food Science & 100 & 100 & 100 & 99 & 100 & 100 & 100 & 100 & 100 & 100 & 100 & 1099 \\
Geography & 179 & 500 & 100 & 498 & 219 & 99 & 169 & 100 & 100 & 171 & 192 & 2327 \\
Health and Medicine & 100 & 499 & 100 & 499 & 100 & 100 & 112 & 100 & 100 & 115 & 107 & 1932 \\
History & 375 & 498 & 100 & 496 & 273 & 100 & 269 & 100 & 100 & 267 & 391 & 2969 \\
Information Technology & 100 & 100 & 100 & 179 & 100 & 100 & 100 & 100 & 100 & 100 & 100 & 1179 \\
International Relations & 100 & 120 & 100 & 257 & 100 & 100 & 100 & 100 & 100 & 100 & 100 & 1277 \\
Language Studies & 100 & 500 & 100 & 391 & 129 & 100 & 149 & 100 & 100 & 210 & 151 & 2030 \\
Law and Ethics & 148 & 500 & 100 & 499 & 157 & 100 & 232 & 100 & 100 & 127 & 160 & 2223 \\
Literature and Linguistics & 111 & 499 & 100 & 495 & 120 & 100 & 113 & 100 & 100 & 154 & 124 & 2016 \\
Logical Reasoning & 407 & 500 & 154 & 499 & 241 & 100 & 359 & 100 & 100 & 251 & 404 & 3115 \\
Materials Science & 100 & 263 & 100 & 458 & 100 & 100 & 100 & 100 & 100 & 100 & 100 & 1621 \\
Media and Communication & 100 & 100 & 100 & 147 & 100 & 100 & 100 & 100 & 100 & 100 & 100 & 1147 \\
Music and Performing Arts & 100 & 100 & 100 & 151 & 100 & 100 & 100 & 100 & 100 & 97 & 100 & 1148 \\
Physics & 500 & 500 & 359 & 500 & 500 & 190 & 500 & 288 & 100 & 499 & 500 & 4436 \\
Politics and Governance & 255 & 499 & 100 & 498 & 247 & 100 & 356 & 100 & 100 & 241 & 424 & 2920 \\
Psychology & 100 & 100 & 100 & 129 & 100 & 100 & 100 & 100 & 100 & 99 & 100 & 1128 \\
Public Administration & 100 & 99 & 99 & 99 & 99 & 100 & 99 & 100 & 99 & 100 & 100 & 1094 \\
Religion and Spirituality & 100 & 100 & 100 & 247 & 100 & 100 & 100 & 100 & 100 & 100 & 100 & 1247 \\
Social Welfare and Development & 100 & 112 & 100 & 193 & 100 & 100 & 100 & 100 & 100 & 99 & 100 & 1204 \\
Sociology & 115 & 500 & 100 & 500 & 100 & 100 & 159 & 100 & 100 & 110 & 169 & 2053 \\
Sports and Recreation & 202 & 500 & 100 & 500 & 178 & 100 & 177 & 100 & 100 & 156 & 210 & 2323 \\
Technology and Innovation & 100 & 500 & 100 & 500 & 100 & 100 & 100 & 100 & 100 & 100 & 100 & 1900 \\
Transportation and Logistics & 100 & 130 & 100 & 317 & 99 & 100 & 98 & 99 & 100 & 100 & 100 & 1343 \\
\midrule
\textbf{TOTAL}  & 6638 & 13536 & 4827 & 14837 & 6234 & 4321 & 6924 & 4525 & 4099 & 6372 & 7304 & \textbf{79617} \\
\bottomrule
\end{tabular}%
}
\caption{Detailed subject level statistics of \milu~across different languages.}
\label{tab:detailed_stats}
\end{table*}
\endgroup

\section{Details about the various models evaluated}
Model details about the different models evaluated in this work is present in Table \ref{tab: models}.

\begingroup
\setlength{\tabcolsep}{5pt} 
\renewcommand{\arraystretch}{1} 
\begin{table*}[t]

\centering
\begin{tabular}{l|cc|r}
\toprule
\textbf{Model Name  } & \textbf{Type} & \textbf{\#Params} & \textbf{Link}\\
\midrule
GPT-4o & Instruct & - & {\color[HTML]{0000EE}\href{https://platform.openai.com/docs/models/gpt-4o}{Link}}  \\
GPT-4o-mini & Instruct & - & {\color[HTML]{0000EE}\href{https://platform.openai.com/docs/models/gpt-4o-mini}{Link}} \\
Gemini-1.5-Pro & Instruct & - & {\color[HTML]{0000EE}\href{https://ai.google.dev/gemini-api/docs/models/gemini?lang=python\#gemini-1.5-pro}{Link}} \\
Gemini-1.5-Flash  & Instruct & - & {\color[HTML]{0000EE}\href{https://ai.google.dev/gemini-api/docs/models/gemini?lang=python\#gemini-1.5-flash}{Link}} \\
Krutrim-spectre-v2 & Instruct & - & {\color[HTML]{0000EE}\href{https://www.olakrutrim.com/}{Link}} \\
\midrule
meta-llama/Llama-3.2-1B & Base & 1B  & {\color[HTML]{0000EE}\href{https://huggingface.co/meta-llama/Llama-3.2-1B}{Link}}  \\
meta-llama/Llama-3.2-1B-Instruct  & Instruct & 1B  & {\color[HTML]{0000EE}\href{https://huggingface.co/meta-llama/Llama-3.2-1B-Instruct}{Link}} \\
\bottomrule
\end{tabular}
\caption{Details about the different models evaluated on \milu.}
\label{tab: models}
\end{table*}
\endgroup

\begingroup
\setlength{\tabcolsep}{12pt} 
\renewcommand{\arraystretch}{1} 
\begin{table*}[]
\centering
\begin{tabular}{lr}
\toprule
\textbf{Subject} & \textbf{ta} \\
\midrule
Agriculture & 29.00/ 28.00/ 26.00 \\
Anthropology & 21.00/ 23.00/ 26.00 \\
Architecture and Design  & 19.00/ 24.00/ 20.00 \\
Arts and Culture & 31.52/ 32.73/ 29.70 \\
Astronomy and Astrophysics & 22.00/ 32.00/ 32.00 \\
Biology  & 24.20/ 19.43/ 24.52 \\
Business and Management  & 36.00/ 28.00/ 24.00 \\
Chemistry & 23.74/ 24.04/ 23.44 \\
Computer Science & 28.49/ 25.00/ 26.74 \\
Defense and Security & 29.00/ 24.00/ 32.00 \\
Earth Sciences & 28.00/ 25.00/ 28.00 \\
Economics & 24.60/ 20.63/ 19.05 \\
Education & 28.00/ 24.00/ 26.00 \\
Energy and Power & 29.00/ 20.00/ 17.00 \\
Engineering & 26.60/ 26.00/ 25.20 \\
Environmental Science & 23.00/ 19.00/ 20.00 \\
Ethics and Human Rights  & 31.00/ 27.00/ 29.00 \\
Finance and Investment & 27.93/ 27.93/ 23.46 \\
Food Science & 27.00/ 21.00/ 21.00 \\
Geography & 22.86/ 26.29/ 24.57 \\
Health and Medicine  & 22.41/ 24.14/ 24.14 \\
History  & 26.18/ 24.73/ 24.00 \\
Information Technology & 28.00/ 23.00/ 25.00 \\
International Relations  & 26.00/ 28.00/ 28.00 \\
Language Studies & 24.72/ 24.34/ 25.47 \\
Law and Ethics & 21.26/ 26.77/ 22.83 \\
Literature and Linguistics & 26.32/ 25.51/ 23.48 \\
Logical Reasoning  & 20.72/ 21.12/ 25.10 \\
Materials Science  & 25.00/ 19.00/ 18.00 \\
Media and Communication  & 18.00/ 19.00/ 19.00 \\
Music and Performing Arts  & 25.00/ 27.00/ 27.00 \\
Physics  & 22.60/ 24.40/ 24.20 \\
Politics and Governance  & 29.05/ 31.12/ 30.29 \\
Psychology & 32.00/ 28.00/ 27.00 \\
Public Administration & 26.00/ 21.00/ 19.00 \\
Religion and Spirituality  & 29.00/ 31.00/ 33.00 \\
Social Welfare and Development & 21.00/ 23.00/ 21.00 \\
Sociology & 19.82/ 22.52/ 23.42 \\
Sports and Recreation & 32.05/ 30.13/ 30.77 \\
Technology and Innovation  & 21.00/ 24.00/ 29.00 \\
Transportation and Logistics & 23.00/ 29.00/ 27.00\\
\bottomrule
\end{tabular}

\caption{Detailed subject-wise evaluation for \textsc{abhinand/tamil-llama-7b-instruct-v0.2} on MILU across different languages. The results reported are X / Y / Z where X denote the 0-shot, Y denotes 1-shot and Z denotes 5-shot performance respectively.}
\end{table*}
\endgroup

\begingroup
\setlength{\tabcolsep}{12pt} 
\renewcommand{\arraystretch}{1} 
\begin{table*}[]
\centering
\begin{tabular}{lr}
\toprule
\textbf{Subject} & \textbf{hi} \\
\midrule
Agriculture & 26.56/ 26.96/ 29.98 \\
Anthropology & 30.30/ 26.26/ 30.30 \\
Architecture and Design  & 28.00/ 31.00/ 26.00 \\
Arts and Culture & 25.45/ 24.65/ 27.45 \\
Astronomy and Astrophysics & 24.22/ 29.69/ 29.69 \\
Biology  & 25.05/ 32.06/ 32.06 \\
Business and Management  & 27.87/ 29.51/ 26.70 \\
Chemistry & 29.66/ 31.46/ 31.26 \\
Computer Science & 28.60/ 30.60/ 30.60 \\
Defense and Security & 28.91/ 22.27/ 24.64 \\
Earth Sciences & 26.05/ 27.45/ 26.25 \\
Economics & 26.81/ 25.20/ 27.02 \\
Education & 27.22/ 25.15/ 26.04 \\
Energy and Power & 31.53/ 29.83/ 31.53 \\
Engineering & 24.00/ 25.60/ 26.40 \\
Environmental Science & 23.65/ 25.85/ 25.65 \\
Ethics and Human Rights  & 33.00/ 25.00/ 29.00 \\
Finance and Investment & 26.60/ 28.20/ 30.20 \\
Food Science & 24.24/ 31.31/ 33.33 \\
Geography & 26.31/ 27.11/ 26.71 \\
Health and Medicine  & 29.06/ 32.87/ 32.67 \\
History  & 27.02/ 28.63/ 27.62 \\
Information Technology & 32.96/ 34.64/ 32.96 \\
International Relations  & 31.91/ 31.91/ 31.52 \\
Language Studies & 25.05/ 28.86/ 25.25 \\
Law and Ethics & 29.20/ 29.40/ 29.20 \\
Literature and Linguistics & 32.87/ 31.86/ 30.06 \\
Logical Reasoning  & 26.45/ 27.45/ 27.25 \\
Materials Science  & 24.67/ 28.60/ 25.55 \\
Media and Communication  & 23.81/ 25.85/ 24.49 \\
Music and Performing Arts  & 25.17/ 24.50/ 25.83 \\
Physics  & 26.20/ 25.20/ 23.40 \\
Politics and Governance  & 28.71/ 27.71/ 26.91 \\
Psychology & 19.38/ 23.26/ 23.26 \\
Public Administration & 39.39/ 41.41/ 40.40 \\
Religion and Spirituality  & 27.82/ 29.03/ 29.44 \\
Social Welfare and Development & 22.28/ 23.32/ 27.98 \\
Sociology & 23.40/ 23.20/ 24.20 \\
Sports and Recreation & 29.40/ 30.20/ 31.60 \\
Technology and Innovation  & 26.20/ 28.60/ 30.60 \\
Transportation and Logistics & 25.87/ 26.50/ 27.13 \\
\bottomrule
\end{tabular}
\caption{Detailed subject-wise evaluation for \textsc{AI4Bharat/Airavata} on MILU across different languages. The results reported are X / Y / Z where X denote the 0-shot, Y denotes 1-shot and Z denotes 5-shot performance respectively.}
\end{table*}
\endgroup

\begingroup
\setlength{\tabcolsep}{12pt} 
\renewcommand{\arraystretch}{1} 
\begin{table*}[]
\centering
\begin{tabular}{lr}
\toprule
\textbf{Subject} & \textbf{hi} \\
\midrule
Agriculture & 27.97/ 27.36/ 28.97 \\
Anthropology & 27.27/ 26.26/ 20.20 \\
Architecture and Design  & 23.00/ 20.00/ 25.00 \\
Arts and Culture & 28.06/ 28.26/ 24.25 \\
Astronomy and Astrophysics & 27.34/ 30.47/ 35.94 \\
Biology  & 31.06/ 30.06/ 30.46 \\
Business and Management  & 26.23/ 24.12/ 29.27 \\
Chemistry & 28.26/ 25.65/ 30.06 \\
Computer Science & 25.80/ 28.80/ 29.80 \\
Defense and Security & 23.22/ 30.81/ 28.91 \\
Earth Sciences & 23.65/ 26.45/ 29.46 \\
Economics & 27.02/ 27.82/ 29.44 \\
Education & 26.33/ 28.99/ 31.07 \\
Energy and Power & 27.80/ 26.44/ 29.83 \\
Engineering & 20.40/ 20.20/ 24.00 \\
Environmental Science & 25.45/ 26.25/ 30.06 \\
Ethics and Human Rights  & 27.00/ 24.00/ 24.00 \\
Finance and Investment & 23.60/ 25.00/ 29.20 \\
Food Science & 32.32/ 37.37/ 28.28 \\
Geography & 26.71/ 25.90/ 28.71 \\
Health and Medicine  & 30.46/ 30.46/ 30.66 \\
History  & 26.01/ 28.02/ 29.03 \\
Information Technology & 27.93/ 31.28/ 36.87 \\
International Relations  & 25.68/ 25.29/ 24.12 \\
Language Studies & 23.05/ 24.85/ 26.65 \\
Law and Ethics & 26.60/ 26.60/ 28.80 \\
Literature and Linguistics & 27.45/ 27.25/ 29.26 \\
Logical Reasoning  & 22.24/ 27.05/ 26.05 \\
Materials Science  & 26.86/ 29.26/ 29.48 \\
Media and Communication  & 23.81/ 26.53/ 32.65 \\
Music and Performing Arts  & 25.17/ 22.52/ 28.48 \\
Physics  & 24.20/ 25.40/ 27.80 \\
Politics and Governance  & 23.09/ 24.70/ 27.11 \\
Psychology & 26.36/ 25.58/ 25.58 \\
Public Administration & 36.36/ 34.34/ 32.32 \\
Religion and Spirituality  & 25.81/ 22.98/ 29.03 \\
Social Welfare and Development & 25.91/ 24.87/ 25.91 \\
Sociology & 26.20/ 25.60/ 24.80 \\
Sports and Recreation & 26.00/ 26.40/ 26.60 \\
Technology and Innovation  & 28.80/ 28.20/ 30.40 \\
Transportation and Logistics & 25.24/ 23.97/ 24.29 \\
\bottomrule
\end{tabular}
\caption{Detailed subject-wise evaluation for \textsc{BhabhaAI/Gajendra-v0.1} on MILU across different languages. The results reported are X / Y / Z where X denote the 0-shot, Y denotes 1-shot and Z denotes 5-shot performance respectively.}
\end{table*}
\endgroup

\begingroup
\setlength{\tabcolsep}{12pt} 
\renewcommand{\arraystretch}{1} 
\begin{table*}[]
\centering
\begin{tabular}{lr}
\toprule
\textbf{Subject} & \textbf{kn} \\
\midrule
Agriculture & 25.00/ 28.00/ 23.00 \\
Anthropology & 25.00/ 26.00/ 26.00 \\
Architecture and Design  & 25.00/ 31.00/ 28.00 \\
Arts and Culture & 26.99/ 29.45/ 24.54 \\
Astronomy and Astrophysics & 28.00/ 30.00/ 30.00 \\
Biology  & 29.10/ 26.23/ 28.28 \\
Business and Management  & 36.00/ 25.00/ 26.00 \\
Chemistry & 31.37/ 25.82/ 28.43 \\
Computer Science & 28.40/ 33.14/ 33.14 \\
Defense and Security & 27.00/ 31.00/ 33.00 \\
Earth Sciences & 29.00/ 31.00/ 33.00 \\
Economics & 29.00/ 32.00/ 37.00 \\
Education & 28.00/ 30.00/ 29.00 \\
Energy and Power & 26.00/ 22.00/ 19.00 \\
Engineering & 24.60/ 27.20/ 28.20 \\
Environmental Science & 21.88/ 23.44/ 29.69 \\
Ethics and Human Rights  & 29.00/ 19.00/ 26.00 \\
Finance and Investment & 29.63/ 28.40/ 31.48 \\
Food Science & 23.00/ 30.00/ 38.00 \\
Geography & 27.85/ 27.40/ 28.77 \\
Health and Medicine  & 23.00/ 29.00/ 24.00 \\
History  & 24.54/ 28.57/ 29.67 \\
Information Technology & 26.00/ 35.00/ 31.00 \\
International Relations  & 32.00/ 36.00/ 26.00 \\
Language Studies & 28.68/ 19.38/ 27.91 \\
Law and Ethics & 25.48/ 22.29/ 26.75 \\
Literature and Linguistics & 18.33/ 23.33/ 23.33 \\
Logical Reasoning  & 21.99/ 21.58/ 25.73 \\
Materials Science  & 28.00/ 24.00/ 23.00 \\
Media and Communication  & 32.00/ 22.00/ 28.00 \\
Music and Performing Arts  & 32.00/ 29.00/ 30.00 \\
Physics  & 26.40/ 26.60/ 26.00 \\
Politics and Governance  & 27.53/ 23.89/ 25.51 \\
Psychology & 23.00/ 34.00/ 31.00 \\
Public Administration & 36.36/ 30.30/ 32.32 \\
Religion and Spirituality  & 27.00/ 27.00/ 31.00 \\
Social Welfare and Development & 28.00/ 26.00/ 32.00 \\
Sociology & 24.00/ 33.00/ 30.00 \\
Sports and Recreation & 22.47/ 29.78/ 27.53 \\
Technology and Innovation  & 25.00/ 29.00/ 32.00 \\
Transportation and Logistics & 25.25/ 26.26/ 29.29 \\
\bottomrule
\end{tabular}
\caption{Detailed subject-wise evaluation for \textsc{Cognitive-Lab/Ambari-7B-base-v0.1} on MILU across different languages. The results reported are X / Y / Z where X denote the 0-shot, Y denotes 1-shot and Z denotes 5-shot performance respectively.}
\end{table*}
\endgroup

\begingroup
\setlength{\tabcolsep}{12pt} 
\renewcommand{\arraystretch}{1} 
\begin{table*}[]
\centering
\begin{tabular}{lr}
\toprule
\textbf{Subject} & \textbf{kn} \\
\midrule
Agriculture & 31.00/ 26.00/ 23.00 \\
Anthropology & 29.00/ 23.00/ 22.00 \\
Architecture and Design  & 29.00/ 21.00/ 23.00 \\
Arts and Culture & 25.15/ 28.83/ 30.06 \\
Astronomy and Astrophysics & 29.00/ 24.00/ 33.00 \\
Biology  & 27.87/ 25.82/ 29.10 \\
Business and Management  & 38.00/ 31.00/ 28.00 \\
Chemistry & 31.37/ 30.07/ 31.05 \\
Computer Science & 24.26/ 25.44/ 30.77 \\
Defense and Security & 27.00/ 22.00/ 19.00 \\
Earth Sciences & 26.00/ 28.00/ 27.00 \\
Economics & 31.00/ 32.00/ 29.00 \\
Education & 32.00/ 30.00/ 32.00 \\
Energy and Power & 26.00/ 21.00/ 26.00 \\
Engineering & 23.20/ 26.20/ 28.20 \\
Environmental Science & 25.78/ 28.12/ 26.56 \\
Ethics and Human Rights  & 29.00/ 25.00/ 28.00 \\
Finance and Investment & 23.46/ 25.31/ 27.78 \\
Food Science & 22.00/ 21.00/ 33.00 \\
Geography & 31.51/ 24.20/ 31.05 \\
Health and Medicine  & 17.00/ 22.00/ 21.00 \\
History  & 27.11/ 28.57/ 30.04 \\
Information Technology & 33.00/ 32.00/ 35.00 \\
International Relations  & 22.00/ 27.00/ 26.00 \\
Language Studies & 25.58/ 30.23/ 23.26 \\
Law and Ethics & 28.03/ 23.57/ 21.66 \\
Literature and Linguistics & 23.33/ 22.50/ 24.17 \\
Logical Reasoning  & 21.16/ 21.16/ 24.07 \\
Materials Science  & 22.00/ 29.00/ 26.00 \\
Media and Communication  & 30.00/ 25.00/ 26.00 \\
Music and Performing Arts  & 29.00/ 27.00/ 25.00 \\
Physics  & 26.40/ 23.60/ 24.20 \\
Politics and Governance  & 22.67/ 29.15/ 25.51 \\
Psychology & 17.00/ 18.00/ 25.00 \\
Public Administration & 24.24/ 23.23/ 28.28 \\
Religion and Spirituality  & 27.00/ 26.00/ 29.00 \\
Social Welfare and Development & 29.00/ 25.00/ 27.00 \\
Sociology & 18.00/ 19.00/ 22.00 \\
Sports and Recreation & 19.10/ 24.16/ 24.16 \\
Technology and Innovation  & 32.00/ 23.00/ 25.00 \\
Transportation and Logistics & 32.32/ 31.31/ 27.27 \\
\bottomrule
\end{tabular}
\caption{Detailed subject-wise evaluation for \textsc{Cognitive-Lab/Ambari-7B-Instruct-v0.1} on MILU across different languages. The results reported are X / Y / Z where X denote the 0-shot, Y denotes 1-shot and Z denotes 5-shot performance respectively.}
\end{table*}
\endgroup

\begingroup
\setlength{\tabcolsep}{12pt} 
\renewcommand{\arraystretch}{1} 
\begin{table*}[]
\centering
\begin{tabular}{lr}
\toprule
\textbf{Subject} & \textbf{hi} \\
\midrule
Agriculture & 29.38/ 30.18/ 36.62 \\
Anthropology & 37.37/ 32.32/ 40.40 \\
Architecture and Design  & 26.00/ 28.00/ 35.00 \\
Arts and Culture & 26.25/ 29.46/ 40.48 \\
Astronomy and Astrophysics & 37.50/ 42.19/ 46.88 \\
Biology  & 35.27/ 35.07/ 50.30 \\
Business and Management  & 31.15/ 32.32/ 38.64 \\
Chemistry & 28.86/ 38.68/ 48.30 \\
Computer Science & 32.60/ 34.00/ 36.60 \\
Defense and Security & 28.91/ 36.02/ 35.55 \\
Earth Sciences & 32.06/ 28.66/ 40.48 \\
Economics & 27.42/ 32.06/ 41.53 \\
Education & 23.96/ 27.51/ 30.77 \\
Energy and Power & 26.10/ 31.19/ 43.39 \\
Engineering & 25.80/ 26.60/ 33.20 \\
Environmental Science & 29.66/ 30.26/ 44.89 \\
Ethics and Human Rights  & 30.00/ 31.00/ 41.00 \\
Finance and Investment & 27.40/ 34.20/ 38.00 \\
Food Science & 24.24/ 33.33/ 47.47 \\
Geography & 31.33/ 30.72/ 42.57 \\
Health and Medicine  & 34.27/ 34.67/ 48.70 \\
History  & 27.02/ 32.86/ 36.90 \\
Information Technology & 34.64/ 37.99/ 50.84 \\
International Relations  & 31.52/ 31.13/ 42.41 \\
Language Studies & 22.65/ 27.05/ 29.26 \\
Law and Ethics & 28.00/ 29.80/ 38.60 \\
Literature and Linguistics & 24.85/ 28.86/ 33.47 \\
Logical Reasoning  & 25.45/ 27.25/ 29.66 \\
Materials Science  & 31.88/ 30.35/ 35.59 \\
Media and Communication  & 25.85/ 26.53/ 40.82 \\
Music and Performing Arts  & 21.19/ 30.46/ 33.77 \\
Physics  & 32.60/ 32.80/ 40.00 \\
Politics and Governance  & 25.50/ 27.71/ 41.97 \\
Psychology & 23.26/ 22.48/ 42.64 \\
Public Administration & 30.30/ 27.27/ 36.36 \\
Religion and Spirituality  & 28.23/ 33.87/ 42.34 \\
Social Welfare and Development & 26.94/ 27.98/ 35.75 \\
Sociology & 23.80/ 26.60/ 28.00 \\
Sports and Recreation & 29.00/ 34.80/ 43.80 \\
Technology and Innovation  & 31.80/ 30.80/ 41.60 \\
Transportation and Logistics & 23.97/ 23.97/ 30.28 \\
\bottomrule
\end{tabular}
\caption{Detailed subject-wise evaluation for \textsc{GenVRadmin/AryaBhatta-GemmaGenZ-Vikas-Merged} on MILU across different languages. The results reported are X / Y / Z where X denote the 0-shot, Y denotes 1-shot and Z denotes 5-shot performance respectively.}
\end{table*}
\endgroup

\begingroup
\setlength{\tabcolsep}{12pt} 
\renewcommand{\arraystretch}{1} 
\begin{table*}[]
\centering
\begin{tabular}{lr}
\toprule
\textbf{Subject} & \textbf{hi} \\
\midrule
Agriculture & 29.38/ 28.97/ 27.16 \\
Anthropology & 36.36/ 36.36/ 28.28 \\
Architecture and Design  & 34.00/ 31.00/ 33.00 \\
Arts and Culture & 34.07/ 33.67/ 34.87 \\
Astronomy and Astrophysics & 42.97/ 41.41/ 37.50 \\
Biology  & 36.67/ 38.08/ 36.07 \\
Business and Management  & 32.55/ 31.15/ 29.04 \\
Chemistry & 37.47/ 34.67/ 31.66 \\
Computer Science & 37.40/ 34.00/ 35.80 \\
Defense and Security & 30.33/ 28.44/ 27.49 \\
Earth Sciences & 31.86/ 31.46/ 33.07 \\
Economics & 34.68/ 33.06/ 30.44 \\
Education & 31.36/ 29.88/ 30.77 \\
Energy and Power & 34.58/ 32.20/ 32.54 \\
Engineering & 28.60/ 27.60/ 27.80 \\
Environmental Science & 31.86/ 31.26/ 30.26 \\
Ethics and Human Rights  & 36.00/ 35.00/ 36.00 \\
Finance and Investment & 29.80/ 30.20/ 28.40 \\
Food Science & 36.36/ 36.36/ 28.28 \\
Geography & 33.33/ 32.73/ 30.12 \\
Health and Medicine  & 44.09/ 40.28/ 38.88 \\
History  & 39.92/ 38.10/ 36.49 \\
Information Technology & 47.49/ 47.49/ 43.02 \\
International Relations  & 40.08/ 40.47/ 34.63 \\
Language Studies & 29.86/ 29.06/ 27.45 \\
Law and Ethics & 41.60/ 39.20/ 40.00 \\
Literature and Linguistics & 29.66/ 30.26/ 29.86 \\
Logical Reasoning  & 30.46/ 28.46/ 29.06 \\
Materials Science  & 29.26/ 31.00/ 30.35 \\
Media and Communication  & 36.73/ 34.01/ 31.29 \\
Music and Performing Arts  & 36.42/ 32.45/ 34.44 \\
Physics  & 33.60/ 29.80/ 29.00 \\
Politics and Governance  & 36.95/ 36.75/ 32.33 \\
Psychology & 41.86/ 37.21/ 33.33 \\
Public Administration & 32.32/ 33.33/ 34.34 \\
Religion and Spirituality  & 29.84/ 31.45/ 29.44 \\
Social Welfare and Development & 30.05/ 30.57/ 30.57 \\
Sociology & 30.00/ 30.00/ 27.40 \\
Sports and Recreation & 32.00/ 30.80/ 28.60 \\
Technology and Innovation  & 36.60/ 36.80/ 34.00 \\
Transportation and Logistics & 30.60/ 29.65/ 28.71 \\
\bottomrule
\end{tabular}
\caption{Detailed subject-wise evaluation for \textsc{manishiitg/open-aditi-v6-llama3} on MILU across different languages. The results reported are X / Y / Z where X denote the 0-shot, Y denotes 1-shot and Z denotes 5-shot performance respectively.}
\end{table*}
\endgroup

\begingroup
\setlength{\tabcolsep}{12pt} 
\renewcommand{\arraystretch}{1} 
\begin{table*}[]
\centering
\begin{tabular}{lr}
\toprule
\textbf{Subject} & \textbf{hi} \\
\midrule
Agriculture & 22.74/ 22.74/ 21.93 \\
Anthropology & 25.25/ 24.24/ 24.24 \\
Architecture and Design  & 32.00/ 28.00/ 29.00 \\
Arts and Culture & 24.85/ 26.05/ 26.25 \\
Astronomy and Astrophysics & 35.94/ 35.94/ 32.03 \\
Biology  & 24.85/ 24.45/ 24.05 \\
Business and Management  & 25.29/ 25.06/ 25.76 \\
Chemistry & 25.85/ 25.25/ 24.45 \\
Computer Science & 26.40/ 27.00/ 25.20 \\
Defense and Security & 27.01/ 24.64/ 25.12 \\
Earth Sciences & 25.45/ 24.05/ 23.85 \\
Economics & 28.02/ 27.42/ 27.42 \\
Education & 26.04/ 24.85/ 24.85 \\
Energy and Power & 23.73/ 24.07/ 25.08 \\
Engineering & 23.60/ 23.40/ 24.00 \\
Environmental Science & 28.86/ 28.26/ 27.86 \\
Ethics and Human Rights  & 30.00/ 31.00/ 22.00 \\
Finance and Investment & 24.00/ 25.00/ 25.40 \\
Food Science & 27.27/ 29.29/ 27.27 \\
Geography & 27.11/ 25.90/ 26.51 \\
Health and Medicine  & 25.45/ 26.05/ 23.85 \\
History  & 24.80/ 25.81/ 26.21 \\
Information Technology & 26.26/ 27.37/ 25.14 \\
International Relations  & 26.85/ 28.40/ 29.96 \\
Language Studies & 21.44/ 21.44/ 22.24 \\
Law and Ethics & 24.80/ 24.80/ 22.80 \\
Literature and Linguistics & 23.65/ 23.85/ 22.24 \\
Logical Reasoning  & 27.05/ 27.86/ 27.45 \\
Materials Science  & 28.17/ 29.04/ 28.60 \\
Media and Communication  & 28.57/ 31.29/ 26.53 \\
Music and Performing Arts  & 21.19/ 22.52/ 23.84 \\
Physics  & 24.40/ 24.00/ 21.60 \\
Politics and Governance  & 26.71/ 25.30/ 22.29 \\
Psychology & 27.91/ 27.91/ 26.36 \\
Public Administration & 26.26/ 27.27/ 28.28 \\
Religion and Spirituality  & 26.21/ 29.44/ 27.82 \\
Social Welfare and Development & 26.94/ 27.98/ 26.42 \\
Sociology & 27.00/ 28.00/ 26.40 \\
Sports and Recreation & 27.80/ 27.40/ 26.60 \\
Technology and Innovation  & 23.80/ 25.20/ 26.40 \\
Transportation and Logistics & 23.97/ 26.81/ 24.92 \\
\bottomrule
\end{tabular}
\caption{Detailed subject-wise evaluation for \textsc{nickmalhotra/projectindus} on MILU across different languages. The results reported are X / Y / Z where X denote the 0-shot, Y denotes 1-shot and Z denotes 5-shot performance respectively.}
\end{table*}
\endgroup

\begingroup
\setlength{\tabcolsep}{12pt} 
\renewcommand{\arraystretch}{1} 
\begin{table*}[]
\centering
\begin{tabular}{lr}
\toprule
\textbf{Subject} & \textbf{hi} \\
\midrule
Agriculture & 24.35/ 30.18/ 29.98 \\
Anthropology & 20.20/ 25.25/ 26.26 \\
Architecture and Design  & 23.00/ 32.00/ 27.00 \\
Arts and Culture & 27.66/ 27.05/ 28.06 \\
Astronomy and Astrophysics & 24.22/ 31.25/ 33.59 \\
Biology  & 27.45/ 26.65/ 28.86 \\
Business and Management  & 26.70/ 25.53/ 29.74 \\
Chemistry & 26.65/ 26.25/ 28.86 \\
Computer Science & 26.80/ 29.20/ 25.80 \\
Defense and Security & 27.96/ 24.64/ 27.96 \\
Earth Sciences & 22.04/ 24.65/ 26.85 \\
Economics & 26.21/ 30.85/ 33.27 \\
Education & 24.26/ 26.92/ 25.15 \\
Energy and Power & 30.17/ 31.19/ 32.54 \\
Engineering & 22.20/ 21.40/ 25.40 \\
Environmental Science & 25.05/ 25.65/ 27.05 \\
Ethics and Human Rights  & 31.00/ 33.00/ 36.00 \\
Finance and Investment & 27.00/ 27.80/ 26.20 \\
Food Science & 20.20/ 24.24/ 30.30 \\
Geography & 24.30/ 27.91/ 27.11 \\
Health and Medicine  & 33.07/ 28.66/ 30.86 \\
History  & 29.44/ 30.24/ 32.66 \\
Information Technology & 25.70/ 30.17/ 35.20 \\
International Relations  & 26.46/ 28.40/ 32.30 \\
Language Studies & 25.45/ 24.85/ 28.46 \\
Law and Ethics & 28.60/ 28.80/ 34.20 \\
Literature and Linguistics & 26.25/ 25.85/ 27.45 \\
Logical Reasoning  & 28.86/ 23.85/ 28.26 \\
Materials Science  & 22.93/ 24.45/ 25.76 \\
Media and Communication  & 29.93/ 27.89/ 24.49 \\
Music and Performing Arts  & 27.15/ 24.50/ 24.50 \\
Physics  & 26.00/ 27.20/ 29.60 \\
Politics and Governance  & 23.49/ 24.90/ 27.91 \\
Psychology & 24.81/ 23.26/ 26.36 \\
Public Administration & 29.29/ 33.33/ 32.32 \\
Religion and Spirituality  & 28.63/ 26.21/ 30.24 \\
Social Welfare and Development & 27.46/ 25.91/ 26.42 \\
Sociology & 28.80/ 28.40/ 25.80 \\
Sports and Recreation & 28.80/ 29.00/ 31.40 \\
Technology and Innovation  & 29.00/ 29.60/ 30.40 \\
Transportation and Logistics & 29.02/ 24.29/ 27.44 \\
\bottomrule
\end{tabular}
\caption{Detailed subject-wise evaluation for \textsc{sarvamai/OpenHathi-7B-Hi-v0.1-Base} on MILU across different languages. The results reported are X / Y / Z where X denote the 0-shot, Y denotes 1-shot and Z denotes 5-shot performance respectively.}
\end{table*}
\endgroup

\begingroup
\setlength{\tabcolsep}{12pt} 
\renewcommand{\arraystretch}{1} 
\begin{table*}[]
\centering
\begin{tabular}{lr}
\toprule
\textbf{Subject} & \textbf{kn} \\
\midrule
Agriculture & 23.00/ 29.00/ 28.00 \\
Anthropology & 30.00/ 41.00/ 38.00 \\
Architecture and Design  & 26.00/ 27.00/ 25.00 \\
Arts and Culture & 26.38/ 26.99/ 29.45 \\
Astronomy and Astrophysics & 26.00/ 30.00/ 21.00 \\
Biology  & 23.77/ 29.51/ 28.69 \\
Business and Management  & 30.00/ 28.00/ 28.00 \\
Chemistry & 28.10/ 27.12/ 25.82 \\
Computer Science & 29.59/ 36.09/ 36.69 \\
Defense and Security & 25.00/ 26.00/ 28.00 \\
Earth Sciences & 24.00/ 24.00/ 24.00 \\
Economics & 30.00/ 32.00/ 29.00 \\
Education & 28.00/ 38.00/ 36.00 \\
Energy and Power & 28.00/ 22.00/ 25.00 \\
Engineering & 23.00/ 23.60/ 24.80 \\
Environmental Science & 23.44/ 32.81/ 30.47 \\
Ethics and Human Rights  & 24.00/ 28.00/ 30.00 \\
Finance and Investment & 26.54/ 31.48/ 29.63 \\
Food Science & 27.00/ 29.00/ 31.00 \\
Geography & 32.88/ 31.96/ 26.94 \\
Health and Medicine  & 17.00/ 27.00/ 30.00 \\
History  & 27.11/ 26.01/ 31.14 \\
Information Technology & 33.00/ 37.00/ 35.00 \\
International Relations  & 25.00/ 25.00/ 29.00 \\
Language Studies & 26.36/ 19.38/ 24.03 \\
Law and Ethics & 26.11/ 26.75/ 24.84 \\
Literature and Linguistics & 22.50/ 25.00/ 21.67 \\
Logical Reasoning  & 22.41/ 23.24/ 22.82 \\
Materials Science  & 25.00/ 26.00/ 24.00 \\
Media and Communication  & 28.00/ 32.00/ 32.00 \\
Music and Performing Arts  & 30.00/ 28.00/ 29.00 \\
Physics  & 26.20/ 28.20/ 26.60 \\
Politics and Governance  & 26.72/ 29.55/ 30.36 \\
Psychology & 29.00/ 29.00/ 25.00 \\
Public Administration & 31.31/ 29.29/ 31.31 \\
Religion and Spirituality  & 26.00/ 27.00/ 35.00 \\
Social Welfare and Development & 32.00/ 34.00/ 31.00 \\
Sociology & 29.00/ 28.00/ 30.00 \\
Sports and Recreation & 26.40/ 30.90/ 29.21 \\
Technology and Innovation  & 29.00/ 30.00/ 34.00 \\
Transportation and Logistics & 36.36/ 31.31/ 32.32 \\
\bottomrule
\end{tabular}
\caption{Detailed subject-wise evaluation for \textsc{Tensoic/Kan-LLaMA-7B-base} on MILU across different languages. The results reported are X / Y / Z where X denote the 0-shot, Y denotes 1-shot and Z denotes 5-shot performance respectively.}
\end{table*}
\endgroup

\begingroup
\setlength{\tabcolsep}{12pt} 
\renewcommand{\arraystretch}{1} 
\begin{table*}[]
\centering
\begin{tabular}{lr}
\toprule
\textbf{Subject} & \textbf{ml} \\
\midrule
Agriculture & 22.22/ 18.18/ 25.25 \\
Anthropology & 24.00/ 20.00/ 21.00 \\
Architecture and Design  & 22.00/ 21.00/ 23.00 \\
Arts and Culture & 25.00/ 20.00/ 23.00 \\
Astronomy and Astrophysics & 27.00/ 28.00/ 26.00 \\
Biology  & 24.24/ 27.27/ 26.26 \\
Business and Management  & 26.00/ 24.00/ 28.00 \\
Chemistry & 25.00/ 26.00/ 26.00 \\
Computer Science & 24.00/ 28.00/ 27.00 \\
Defense and Security & 18.00/ 18.00/ 18.00 \\
Earth Sciences & 30.00/ 31.00/ 34.00 \\
Economics & 23.00/ 26.00/ 27.00 \\
Education & 23.00/ 21.00/ 22.00 \\
Energy and Power & 32.00/ 28.00/ 30.00 \\
Engineering & 27.35/ 27.78/ 25.64 \\
Environmental Science & 30.00/ 31.00/ 28.00 \\
Ethics and Human Rights  & 29.00/ 19.00/ 24.00 \\
Finance and Investment & 35.00/ 31.00/ 35.00 \\
Food Science & 27.00/ 22.00/ 20.00 \\
Geography & 30.30/ 27.27/ 24.24 \\
Health and Medicine  & 23.00/ 23.00/ 20.00 \\
History  & 30.00/ 22.00/ 25.00 \\
Information Technology & 35.00/ 37.00/ 37.00 \\
International Relations  & 23.00/ 25.00/ 24.00 \\
Language Studies & 21.00/ 17.00/ 20.00 \\
Law and Ethics & 27.00/ 30.00/ 27.00 \\
Literature and Linguistics & 28.00/ 24.00/ 25.00 \\
Logical Reasoning  & 29.00/ 26.00/ 27.00 \\
Materials Science  & 28.00/ 25.00/ 20.00 \\
Media and Communication  & 24.00/ 24.00/ 22.00 \\
Music and Performing Arts  & 25.00/ 21.00/ 24.00 \\
Physics  & 25.79/ 27.89/ 30.53 \\
Politics and Governance  & 30.00/ 28.00/ 31.00 \\
Psychology & 26.00/ 25.00/ 24.00 \\
Public Administration & 23.00/ 25.00/ 20.00 \\
Religion and Spirituality  & 27.00/ 24.00/ 26.00 \\
Social Welfare and Development & 34.00/ 33.00/ 30.00 \\
Sociology & 20.00/ 21.00/ 20.00 \\
Sports and Recreation & 31.00/ 27.00/ 25.00 \\
Technology and Innovation  & 25.00/ 26.00/ 26.00 \\
Transportation and Logistics & 24.00/ 25.00/ 32.00 \\
\bottomrule
\end{tabular}
\caption{Detailed subject-wise evaluation for \textsc{abhinand/malayalam-llama-7b-instruct-v0.1} on MILU across different languages. The results reported are X / Y / Z where X denote the 0-shot, Y denotes 1-shot and Z denotes 5-shot performance respectively.}
\end{table*}
\endgroup

\begingroup
\setlength{\tabcolsep}{12pt} 
\renewcommand{\arraystretch}{1} 
\begin{table*}[]
\centering
\begin{tabular}{lr}
\toprule
\textbf{Subject} & \textbf{te} \\
\midrule
Agriculture & 25.00/ 23.00/ 21.00 \\
Anthropology & 33.00/ 30.00/ 33.00 \\
Architecture and Design  & 24.00/ 20.00/ 20.00 \\
Arts and Culture & 28.64/ 29.65/ 28.14 \\
Astronomy and Astrophysics & 25.00/ 25.00/ 25.00 \\
Biology  & 24.53/ 23.27/ 22.01 \\
Business and Management  & 36.79/ 36.79/ 39.62 \\
Chemistry & 25.69/ 27.43/ 27.93 \\
Computer Science & 25.56/ 30.04/ 28.70 \\
Defense and Security & 26.00/ 25.00/ 25.00 \\
Earth Sciences & 27.00/ 29.00/ 29.00 \\
Economics & 24.55/ 28.74/ 29.94 \\
Education & 22.00/ 22.00/ 28.00 \\
Energy and Power & 26.00/ 26.00/ 24.00 \\
Engineering & 30.40/ 29.20/ 28.00 \\
Environmental Science & 21.23/ 23.46/ 23.46 \\
Ethics and Human Rights  & 21.00/ 22.00/ 20.00 \\
Finance and Investment & 26.79/ 24.64/ 26.79 \\
Food Science & 28.00/ 25.00/ 21.00 \\
Geography & 28.65/ 28.12/ 23.44 \\
Health and Medicine  & 28.04/ 30.84/ 28.04 \\
History  & 31.71/ 32.48/ 29.67 \\
Information Technology & 30.00/ 38.00/ 33.00 \\
International Relations  & 27.00/ 29.00/ 31.00 \\
Language Studies & 27.01/ 28.16/ 27.59 \\
Law and Ethics & 27.50/ 24.37/ 26.87 \\
Literature and Linguistics & 25.35/ 21.83/ 18.31 \\
Logical Reasoning  & 24.75/ 24.50/ 25.25 \\
Materials Science  & 26.00/ 24.00/ 27.00 \\
Media and Communication  & 23.00/ 23.00/ 25.00 \\
Music and Performing Arts  & 28.00/ 23.00/ 30.00 \\
Physics  & 22.80/ 25.20/ 28.20 \\
Politics and Governance  & 27.59/ 27.59/ 29.48 \\
Psychology & 28.00/ 22.00/ 22.00 \\
Public Administration & 35.00/ 31.00/ 28.00 \\
Religion and Spirituality  & 22.00/ 24.00/ 24.00 \\
Social Welfare and Development & 30.00/ 27.00/ 29.00 \\
Sociology & 21.18/ 20.59/ 18.82 \\
Sports and Recreation & 24.76/ 25.71/ 26.67 \\
Technology and Innovation  & 29.00/ 24.00/ 22.00 \\
Transportation and Logistics & 26.00/ 22.00/ 26.00 \\
\bottomrule
\end{tabular}
\caption{Detailed subject-wise evaluation for \textsc{abhinand/telugu-llama-7b-instruct-v0.1} on MILU across different languages. The results reported are X / Y / Z where X denote the 0-shot, Y denotes 1-shot and Z denotes 5-shot performance respectively.}
\end{table*}
\endgroup

\begingroup
\setlength{\tabcolsep}{12pt} 
\renewcommand{\arraystretch}{1} 
\begin{table*}[]
\centering
\begin{tabular}{lr}
\toprule
\textbf{Subject} & \textbf{mr} \\
\midrule
Agriculture & 20.00/ 20.00/ 17.00 \\
Anthropology & 15.00/ 20.00/ 20.00 \\
Architecture and Design  & 18.00/ 19.00/ 19.00 \\
Arts and Culture & 20.22/ 20.77/ 22.95 \\
Astronomy and Astrophysics & 20.00/ 19.00/ 19.00 \\
Biology  & 24.18/ 23.58/ 23.28 \\
Business and Management  & 20.00/ 21.00/ 22.00 \\
Chemistry & 23.75/ 25.33/ 25.59 \\
Computer Science & 26.63/ 25.44/ 27.22 \\
Defense and Security & 24.00/ 22.00/ 19.00 \\
Earth Sciences & 23.00/ 29.00/ 28.00 \\
Economics & 28.17/ 26.76/ 28.17 \\
Education & 24.20/ 23.57/ 22.29 \\
Energy and Power & 27.00/ 25.00/ 22.00 \\
Engineering & 21.60/ 21.60/ 22.20 \\
Environmental Science & 23.16/ 23.16/ 26.84 \\
Ethics and Human Rights  & 20.00/ 18.00/ 17.00 \\
Finance and Investment & 23.16/ 24.29/ 24.29 \\
Food Science & 25.00/ 27.00/ 21.00 \\
Geography & 21.30/ 20.71/ 18.93 \\
Health and Medicine  & 20.54/ 27.68/ 23.21 \\
History  & 24.91/ 24.54/ 25.65 \\
Information Technology & 25.00/ 24.00/ 23.00 \\
International Relations  & 24.00/ 27.00/ 23.00 \\
Language Studies & 21.48/ 23.49/ 23.49 \\
Law and Ethics & 21.12/ 21.55/ 23.71 \\
Literature and Linguistics & 20.35/ 23.01/ 22.12 \\
Logical Reasoning  & 26.18/ 24.79/ 26.74 \\
Materials Science  & 16.00/ 19.00/ 16.00 \\
Media and Communication  & 20.00/ 17.00/ 17.00 \\
Music and Performing Arts  & 27.00/ 27.00/ 31.00 \\
Physics  & 25.00/ 24.40/ 23.40 \\
Politics and Governance  & 21.07/ 20.22/ 22.75 \\
Psychology & 24.00/ 25.00/ 24.00 \\
Public Administration & 24.24/ 25.25/ 18.18 \\
Religion and Spirituality  & 19.00/ 22.00/ 24.00 \\
Social Welfare and Development & 18.00/ 19.00/ 22.00 \\
Sociology & 18.24/ 19.50/ 21.38 \\
Sports and Recreation & 25.99/ 22.03/ 22.03 \\
Technology and Innovation  & 18.00/ 20.00/ 18.00 \\
Transportation and Logistics & 21.43/ 26.53/ 24.49 \\
\bottomrule
\end{tabular}
\caption{Detailed subject-wise evaluation for \textsc{smallstepai/Misal-7B-instruct-v0.1} on MILU across different languages. The results reported are X / Y / Z where X denote the 0-shot, Y denotes 1-shot and Z denotes 5-shot performance respectively.}
\end{table*}
\endgroup

\begingroup
\setlength{\tabcolsep}{12pt} 
\renewcommand{\arraystretch}{1} 
\begin{table*}[]
\centering
\begin{tabular}{lr}
\toprule
\textbf{Subject} & \textbf{mr} \\
\midrule
Agriculture & 34.00/ 28.00/ 26.00 \\
Anthropology & 27.00/ 35.00/ 37.00 \\
Architecture and Design  & 21.00/ 18.00/ 19.00  \\
Arts and Culture & 25.68/ 25.14/ 26.23 \\
Astronomy and Astrophysics & 25.00/ 23.00/ 24.00  \\
Biology  & 23.88/ 25.97/ 25.07 \\
Business and Management  & 19.00/ 27.00/ 26.00 \\
Chemistry & 29.82/ 27.70/ 29.02 \\
Computer Science & 20.12/ 24.26/ 23.08  \\
Defense and Security & 23.00/ 26.00/ 22.00  \\
Earth Sciences & 30.00/ 28.00/ 25.00  \\
Economics & 27.46/ 29.58/ 30.99  \\
Education & 38.22/ 28.66/ 28.66  \\
Energy and Power & 25.00/ 24.00/ 28.00  \\
Engineering & 21.60/ 28.00/ 26.80  \\
Environmental Science & 26.32/ 30.53/ 32.11  \\
Ethics and Human Rights  & 20.00/ 22.00/ 25.00  \\
Finance and Investment & 25.99/ 22.03/ 24.29  \\
Food Science & 19.00/ 17.00/ 21.00  \\
Geography & 24.85/ 21.30/ 23.67  \\
Health and Medicine  & 29.46/ 33.93/ 26.79  \\
History  & 26.77/ 27.14/ 28.25  \\
Information Technology & 19.00/ 17.00/ 21.00  \\
International Relations  & 30.00/ 30.00/ 25.00  \\
Language Studies & 22.15/ 22.15/ 20.81  \\
Law and Ethics & 25.00/ 29.31/ 32.33  \\
Literature and Linguistics & 22.12/ 27.43/ 29.20  \\
Logical Reasoning  & 30.64/ 30.64/ 30.36  \\
Materials Science  & 37.00/ 27.00/ 24.00  \\
Media and Communication  & 24.00/ 31.00/ 31.00  \\
Music and Performing Arts  & 23.00/ 23.00/ 30.00  \\
Physics  & 26.00/ 24.00/ 23.00  \\
Politics and Governance  & 30.62/ 25.84/ 26.12  \\
Psychology & 30.00/ 22.00/ 21.00  \\
Public Administration & 29.29/ 25.25/ 25.25  \\
Religion and Spirituality  & 18.00/ 26.00/ 28.00  \\
Social Welfare and Development & 25.00/ 18.00/ 23.00  \\
Sociology & 25.79/ 19.50/ 16.35  \\
Sports and Recreation & 26.55/ 24.29/ 29.38  \\
Technology and Innovation  & 28.00/ 30.00/ 31.00  \\
Transportation and Logistics & 27.55/ 25.51/ 28.57  \\
\bottomrule
\end{tabular}
\caption{Detailed subject-wise evaluation for \textsc{smallstepai/Misal-7B-base-v0.1} on MILU across different languages. The results reported are X / Y / Z where X denote the 0-shot, Y denotes 1-shot and Z denotes 5-shot performance respectively.}
\end{table*}
\endgroup

\begingroup
\setlength{\tabcolsep}{12pt} 
\renewcommand{\arraystretch}{1} 
\begin{table*}[]
\centering
\begin{tabular}{lr}
\toprule
\textbf{Subject} & \textbf{te} \\
\midrule
Agriculture & 25.00/ 21.00/ 29.00 \\
Anthropology & 22.00/ 19.00/ 21.00 \\
Architecture and Design  & 19.00/ 14.00/ 17.00 \\
Arts and Culture & 24.62/ 21.11/ 22.11 \\
Astronomy and Astrophysics & 31.00/ 27.00/ 28.00 \\
Biology  & 19.81/ 19.50/ 22.33 \\
Business and Management  & 17.92/ 16.98/ 18.87 \\
Chemistry & 27.93/ 27.43/ 29.43 \\
Computer Science & 27.80/ 27.35/ 30.49 \\
Defense and Security & 35.00/ 29.00/ 32.00 \\
Earth Sciences & 31.00/ 32.00/ 31.00 \\
Economics & 26.95/ 23.35/ 23.35 \\
Education & 30.00/ 24.00/ 23.00 \\
Energy and Power & 24.00/ 20.00/ 26.00 \\
Engineering & 26.80/ 27.00/ 26.60 \\
Environmental Science & 26.82/ 23.46/ 24.02 \\
Ethics and Human Rights  & 17.00/ 18.00/ 19.00 \\
Finance and Investment & 25.00/ 22.50/ 24.29 \\
Food Science & 22.00/ 24.00/ 25.00 \\
Geography & 26.56/ 26.56/ 27.08 \\
Health and Medicine  & 25.23/ 28.04/ 29.91 \\
History  & 29.16/ 26.85/ 26.09 \\
Information Technology & 32.00/ 30.00/ 33.00 \\
International Relations  & 16.00/ 21.00/ 21.00 \\
Language Studies & 28.74/ 27.59/ 29.31 \\
Law and Ethics & 27.50/ 20.62/ 23.75 \\
Literature and Linguistics & 28.17/ 23.94/ 25.35 \\
Logical Reasoning  & 26.49/ 26.73/ 26.24 \\
Materials Science  & 24.00/ 27.00/ 31.00 \\
Media and Communication  & 33.00/ 36.00/ 34.00 \\
Music and Performing Arts  & 23.00/ 17.00/ 25.00 \\
Physics  & 26.80/ 25.40/ 27.40 \\
Politics and Governance  & 29.25/ 28.77/ 27.36 \\
Psychology & 26.00/ 26.00/ 23.00 \\
Public Administration & 27.00/ 32.00/ 34.00 \\
Religion and Spirituality  & 20.00/ 27.00/ 26.00 \\
Social Welfare and Development & 32.00/ 27.00/ 30.00 \\
Sociology & 25.29/ 24.71/ 25.88 \\
Sports and Recreation & 21.43/ 22.38/ 21.90 \\
Technology and Innovation  & 22.00/ 21.00/ 21.00 \\
Transportation and Logistics & 22.00/ 24.00/ 27.00 \\
\bottomrule
\end{tabular}
\caption{Detailed subject-wise evaluation for \textsc{Telugu-LLM-Labs/Telugu-Llama2-7B-v0-Base} on MILU across different languages. The results reported are X / Y / Z where X denote the 0-shot, Y denotes 1-shot and Z denotes 5-shot performance respectively.}
\end{table*}
\endgroup

\begingroup
\setlength{\tabcolsep}{3pt} 
\renewcommand{\arraystretch}{1} 
\begin{table*}[t]
\centering
\resizebox{\textwidth}{!}{%
\begin{tabular}{lcccccccccccc}
\toprule
\textbf{Topic} & \multicolumn{1}{c}{\textit{\textbf{bn}}} & \multicolumn{1}{c}{\textit{\textbf{en}}} & \multicolumn{1}{c}{\textit{\textbf{gu}}} & \multicolumn{1}{c}{\textit{\textbf{hi}}} & \multicolumn{1}{c}{\textbf{\textit{kn}}} & \multicolumn{1}{c}{\textbf{\textit{ml}}} & \multicolumn{1}{c}{\textbf{\textit{mr}}} & \multicolumn{1}{c}{\textbf{\textit{or}}} & \multicolumn{1}{c}{\textbf{\textit{pa}}} & \multicolumn{1}{c}{\textbf{\textit{ta}}} & \multicolumn{1}{c}{\textbf{\textit{te}}} \\
\midrule
Agriculture                    & 70   & 79.5 & 79.8 & 75.7 & 74   & 63.6 & 67   & 63   & 74   & 65   & 72   \\
Anthropology                   & 79   & 80   & 69   & 76.8 & 73   & 73   & 83   & 72   & 72   & 76   & 80   \\
Architecture and Design        & 77   & 80   & 80   & 77   & 70   & 75   & 72   & 74   & 69   & 66   & 81   \\
Arts and Culture               & 86.1 & 85.8 & 83   & 82   & 86.5 & 80   & 83.6 & 78   & 76   & 74.5 & 82.9 \\
Astronomy and Astrophysics     & 88   & 95   & 92   & 88.3 & 88   & 85   & 84   & 87   & 85   & 88   & 88   \\
Biology                        & 87.3 & 92.4 & 83   & 90   & 82   & 82.8 & 79.1 & 82   & 89   & 76.8 & 81.4 \\
Business and Management        & 67   & 73.4 & 62   & 73.8 & 64   & 61   & 67   & 65   & 67   & 63   & 59.4 \\
\bottomrule
\end{tabular}%
}
\caption{Detailed subject-wise evaluation for \textsc{GPT-4o} on MILU across
different languages. The results reported are for 0-shot experiments.}
\end{table*}
\endgroup

\begingroup
\setlength{\tabcolsep}{3pt} 
\renewcommand{\arraystretch}{1} 
\begin{table*}[t]
\centering
\resizebox{\textwidth}{!}{%
\begin{tabular}{lcccccccccccc}
\toprule
\textbf{Topic} & \multicolumn{1}{c}{\textit{\textbf{bn}}} & \multicolumn{1}{c}{\textit{\textbf{en}}} & \multicolumn{1}{c}{\textit{\textbf{gu}}} & \multicolumn{1}{c}{\textit{\textbf{hi}}} & \multicolumn{1}{c}{\textbf{\textit{kn}}} & \multicolumn{1}{c}{\textbf{\textit{ml}}} & \multicolumn{1}{c}{\textbf{\textit{mr}}} & \multicolumn{1}{c}{\textbf{\textit{or}}} & \multicolumn{1}{c}{\textbf{\textit{pa}}} & \multicolumn{1}{c}{\textbf{\textit{ta}}} & \multicolumn{1}{c}{\textbf{\textit{te}}} \\
\midrule
Agriculture                    & 60   & 64   & 60.6 & 61   & 56   & 43.4 & 45   & 43   & 51   & 51   & 53   \\
Anthropology                   & 55   & 67   & 49   & 63.6 & 48   & 49   & 60   & 39   & 58   & 52   & 60   \\
Architecture and Design        & 55   & 60   & 52   & 53   & 48   & 50   & 53   & 39   & 51   & 51   & 56   \\
Arts and Culture               & 65.5 & 76   & 54   & 65.9 & 57.7 & 57   & 57.9 & 45   & 62   & 53.9 & 55.8 \\
Astronomy and Astrophysics     & 76   & 92   & 78   & 76.6 & 73   & 68   & 68   & 62   & 67   & 67   & 71   \\
Biology                        & 75   & 86.8 & 68   & 79.6 & 57.8 & 59.6 & 57.9 & 49   & 68   & 53.8 & 58.8 \\
Business and Management        & 62   & 63.8 & 56   & 60.2 & 58   & 46   & 54   & 44   & 57   & 49   & 44.3 \\
\bottomrule
\end{tabular}%
}
\caption{Detailed subject-wise evaluation for \textsc{GPT-4o-mini} on MILU across
different languages. The results reported are for 0-shot experiments.}
\end{table*}
\endgroup

\begingroup
\setlength{\tabcolsep}{3pt} 
\renewcommand{\arraystretch}{1} 
\begin{table*}[t]
\centering
\resizebox{\textwidth}{!}{%
\begin{tabular}{lcccccccccccc}
\toprule
\textbf{Topic} & \multicolumn{1}{c}{\textit{\textbf{bn}}} & \multicolumn{1}{c}{\textit{\textbf{en}}} & \multicolumn{1}{c}{\textit{\textbf{gu}}} & \multicolumn{1}{c}{\textit{\textbf{hi}}} & \multicolumn{1}{c}{\textbf{\textit{kn}}} & \multicolumn{1}{c}{\textbf{\textit{ml}}} & \multicolumn{1}{c}{\textbf{\textit{mr}}} & \multicolumn{1}{c}{\textbf{\textit{or}}} & \multicolumn{1}{c}{\textbf{\textit{pa}}} & \multicolumn{1}{c}{\textbf{\textit{ta}}} & \multicolumn{1}{c}{\textbf{\textit{te}}} \\
\midrule
Agriculture                    & 60   & 76.7 & 66.7 & 72.2 & 66   & 51.5 & 53   & 52   & 59   & 50   & 60   \\
Anthropology                   & 67   & 73   & 57   & 68.7 & 60   & 49   & 69   & 50   & 58   & 60   & 69   \\
Architecture and Design        & 59   & 69   & 46   & 65   & 48   & 53   & 67   & 51   & 52   & 57   & 57   \\
Arts and Culture               & 58.8 & 77.8 & 58   & 69.9 & 61.3 & 51   & 62.3 & 51   & 56   & 56.4 & 58.3 \\
Astronomy and Astrophysics     & 81   & 94   & 89   & 82.8 & 82   & 73   & 81   & 81   & 74   & 78   & 74   \\
Biology                        & 80.3 & 90.2 & 73   & 85.8 & 83.2 & 77.8 & 71.9 & 70   & 81   & 74.8 & 70.1 \\
Business and Management        & 61   & 71.8 & 56   & 67.9 & 61   & 62   & 61   & 44   & 62   & 60   & 51.9 \\
\bottomrule
\end{tabular}%
}
\caption{Detailed subject-wise evaluation for  \textsc{Gemini-1.5-Pro} on MILU across
different languages. The results reported are for 0-shot experiments.}
\end{table*}
\endgroup

\begingroup
\setlength{\tabcolsep}{3pt} 
\renewcommand{\arraystretch}{1} 
\begin{table*}[t]
\centering
\resizebox{\textwidth}{!}{%
\begin{tabular}{lcccccccccccc}
\toprule
\textbf{Topic} & \multicolumn{1}{c}{\textit{\textbf{bn}}} & \multicolumn{1}{c}{\textit{\textbf{en}}} & \multicolumn{1}{c}{\textit{\textbf{gu}}} & \multicolumn{1}{c}{\textit{\textbf{hi}}} & \multicolumn{1}{c}{\textbf{\textit{kn}}} & \multicolumn{1}{c}{\textbf{\textit{ml}}} & \multicolumn{1}{c}{\textbf{\textit{mr}}} & \multicolumn{1}{c}{\textbf{\textit{or}}} & \multicolumn{1}{c}{\textbf{\textit{pa}}} & \multicolumn{1}{c}{\textbf{\textit{ta}}} & \multicolumn{1}{c}{\textbf{\textit{te}}} \\
\midrule
Agriculture                    & 64   & 65.6 & 58.6 & 62.6 & 60   & 55.6 & 57   & 59   & 57   & 52   & 63   \\
Anthropology                   & 58   & 66   & 49   & 57.6 & 58   & 54   & 62   & 59   & 65   & 54   & 56   \\
Architecture and Design        & 58   & 60   & 57   & 60   & 51   & 52   & 64   & 46   & 58   & 52   & 51   \\
Arts and Culture               & 58.2 & 71   & 60   & 63.9 & 62.6 & 54   & 59.6 & 46   & 54   & 60   & 54.8 \\
Astronomy and Astrophysics     & 79   & 91   & 84   & 83.6 & 78   & 71   & 83   & 80   & 76   & 80   & 71   \\
Biology                        & 75.4 & 87   & 72   & 84.6 & 78.6 & 72.7 & 69.3 & 66   & 82   & 69.7 & 73   \\
Business and Management        & 50   & 64.7 & 46   & 59.3 & 56   & 53   & 51   & 56   & 62   & 52   & 46.2 \\
\bottomrule
\end{tabular}%
}
\caption{Detailed subject-wise evaluation for \textsc{Gemini-1.5-Flash} on MILU across
different languages. The results reported are for 0-shot experiments.}
\end{table*}
\endgroup

\begingroup
\setlength{\tabcolsep}{3pt} 
\renewcommand{\arraystretch}{1} 
\begin{table*}[t]
\centering
\resizebox{\textwidth}{!}{%
\begin{tabular}{lcccccccccccc}
\toprule
\textbf{Topic} & \multicolumn{1}{c}{\textit{\textbf{bn}}} & \multicolumn{1}{c}{\textit{\textbf{en}}} & \multicolumn{1}{c}{\textit{\textbf{gu}}} & \multicolumn{1}{c}{\textit{\textbf{hi}}} & \multicolumn{1}{c}{\textbf{\textit{kn}}} & \multicolumn{1}{c}{\textbf{\textit{ml}}} & \multicolumn{1}{c}{\textbf{\textit{mr}}} & \multicolumn{1}{c}{\textbf{\textit{or}}} & \multicolumn{1}{c}{\textbf{\textit{pa}}} & \multicolumn{1}{c}{\textbf{\textit{ta}}} & \multicolumn{1}{c}{\textbf{\textit{te}}} \\
\midrule
Agriculture                    & 51   & 45.3 & 52   & 42.7 & 49   & 36.7 & 38   & 38   & 37.9 & 42   & 37   \\
Anthropology                   & 47   & 42   & 43   & 43.4 & 41   & 40.2 & 44   & 38   & 40   & 36   & 35   \\
Architecture and Design        & 38   & 50   & 41.9 & 43   & 45   & 42   & 49.5 & 32.3 & 34   & 34   & 43.4 \\
Arts and Culture               & 55.8 & 57.4 & 54   & 54.8 & 52.8 & 52   & 46.1 & 37   & 44   & 48.8 & 47.7 \\
Astronomy and Astrophysics     & 53   & 53   & 57   & 55.9 & 53   & 50   & 57.6 & 43   & 37   & 51   & 45   \\
Biology                        & 57.4 & 63.7 & 61   & 57.6 & 60.7 & 45.5 & 47.6 & 40   & 48   & 48.1 & 46.1 \\
Business and Management        & 40   & 41.2 & 46   & 43.7 & 43   & 38   & 42.9 & 35   & 41   & 36   & 29.2 \\
\bottomrule
\end{tabular}%
}
\caption{Detailed subject-wise evaluation for \textsc{Krutrim-spectre-v2} on MILU across
different languages. The results reported are for 0-shot experiments.}
\end{table*}
\endgroup

\begingroup
\setlength{\tabcolsep}{3pt} 
\renewcommand{\arraystretch}{1} 
\begin{table*}[t]
\centering
\resizebox{\textwidth}{!}{%
\begin{tabular}{lcccccccccccc}
\toprule
\textbf{Topic} & \multicolumn{1}{c}{\textit{\textbf{bn}}} & \multicolumn{1}{c}{\textit{\textbf{en}}} & \multicolumn{1}{c}{\textit{\textbf{gu}}} & \multicolumn{1}{c}{\textit{\textbf{hi}}} & \multicolumn{1}{c}{\textbf{\textit{kn}}} & \multicolumn{1}{c}{\textbf{\textit{ml}}} & \multicolumn{1}{c}{\textbf{\textit{mr}}} & \multicolumn{1}{c}{\textbf{\textit{or}}} & \multicolumn{1}{c}{\textbf{\textit{pa}}} & \multicolumn{1}{c}{\textbf{\textit{ta}}} & \multicolumn{1}{c}{\textbf{\textit{te}}} \\
\midrule
Agriculture & 32 & 35.4 & 38.4 & 31 & 30 & 33.3 & 27 & 31 & 30 & 28 & 28 \\
Anthropology & 41 & 40 & 27 & 38.4 & 31 & 27 & 31 & 29 & 34 & 26 & 35 \\
Architecture and Design & 31 & 35 & 32 & 30 & 30 & 20 & 19 & 30 & 31 & 30 & 34 \\
Arts and Culture & 28.5 & 37.2 & 29 & 31.3 & 32.5 & 24 & 31.1 & 27 & 23 & 29.7 & 30.1 \\
Astronomy and Astrophysics & 46 & 45 & 42 & 42.2 & 39 & 24 & 41 & 31 & 37 & 40 & 37 \\
Biology & 40.5 & 46.9 & 36 & 38.7 & 38.9 & 30.3 & 31.3 & 30 & 39 & 31.2 & 34.9 \\
Business and Management & 31 & 37.8 & 30 & 34 & 27 & 27 & 29 & 25 & 28 & 34 & 25.5 \\
\bottomrule
\end{tabular}%
}
\caption{Detailed subject-wise evaluation for \textsc{sarvamai/sarvam-1} on MILU across
different languages. The results reported are for 5-shot experiments.}
\end{table*}
\endgroup

\begingroup
\setlength{\tabcolsep}{10pt} 
\renewcommand{\arraystretch}{1.3} 

\begin{table*}[]

\centering
\resizebox{\textwidth}{!}{
\begin{tabular}{lcccccccccccc}
\toprule
\textbf{Topic} & \multicolumn{1}{c}{\textit{\textbf{bn}}} & \multicolumn{1}{c}{\textit{\textbf{en}}} & \multicolumn{1}{c}{\textit{\textbf{gu}}} & \multicolumn{1}{c}{\textit{\textbf{hi}}} & \multicolumn{1}{c}{\textbf{\textit{kn}}} & \multicolumn{1}{c}{\textbf{\textit{ml}}} & \multicolumn{1}{c}{\textbf{\textit{mr}}} & \multicolumn{1}{c}{\textbf{\textit{or}}} & \multicolumn{1}{c}{\textbf{\textit{pa}}} & \multicolumn{1}{c}{\textbf{\textit{ta}}} & \multicolumn{1}{c}{\textbf{\textit{te}}} \\
\midrule
Agriculture & 28.0/ 34.0/ 29.0 & 24.3/ 29.0/ 29.8 & 29.3/ 30.3/ 23.2 & 22.1/ 26.8/ 26.0 & 21.0/ 27.0/ 26.0 & 38.4/ 31.3/ 31.3 & 27.0/ 26.0/ 27.0 & 29.0/ 26.0/ 30.0 & 19.0/ 26.0/ 25.0 & 26.0/ 29.0/ 26.0 & 26.0/ 25.0/ 26.0 \\
Anthropology & 21.0/ 15.0/ 25.0 & 24.0/ 37.0/ 39.0 & 21.0/ 23.0/ 28.0 & 21.2/ 24.2/ 28.3 & 31.0/ 27.0/ 28.0 & 25.0/ 25.0/ 29.0 & 23.0/ 25.0/ 21.0 & 28.0/ 29.0/ 28.0 & 19.0/ 20.0/ 23.0 & 27.0/ 23.0/ 23.0 & 19.0/ 24.0/ 19.0 \\
Architecture and Design & 25.0/ 19.0/ 25.0 & 27.0/ 23.0/ 39.0 & 30.0/ 25.0/ 20.0 & 22.0/ 29.0/ 38.0 & 28.0/ 24.0/ 22.0 & 28.0/ 27.0/ 25.0 & 23.0/ 24.0/ 25.0 & 17.0/ 23.0/ 22.0 & 25.0/ 23.0/ 27.0 & 29.0/ 32.0/ 29.0 & 28.0/ 26.0/ 24.0 \\
Arts and Culture & 24.2/ 24.9/ 29.7 & 31.0/ 32.8/ 36.8 & 30.0/ 24.0/ 22.0 & 23.2/ 23.6/ 27.1 & 23.3/ 20.9/ 24.5 & 31.0/ 27.0/ 27.0 & 21.9/ 24.6/ 24.0 & 26.0/ 24.0/ 21.0 & 29.0/ 25.0/ 27.0 & 29.7/ 26.1/ 23.6 & 27.1/ 28.1/ 23.1 \\
Astronomy and Astrophysics & 22.0/ 24.0/ 20.0 & 27.0/ 37.0/ 35.0 & 30.0/ 34.0/ 33.0 & 24.2/ 28.9/ 28.9 & 24.0/ 22.0/ 23.0 & 24.0/ 24.0/ 23.0 & 24.0/ 27.0/ 35.0 & 25.0/ 30.0/ 29.0 & 24.0/ 23.0/ 24.0 & 21.0/ 20.0/ 16.0 & 20.0/ 31.0/ 19.0 \\
Biology & 23.6/ 24.3/ 21.5 & 33.7/ 40.7/ 44.7 & 21.0/ 17.0/ 21.0 & 26.1/ 28.3/ 30.7 & 23.8/ 28.3/ 25.8 & 24.2/ 22.2/ 24.2 & 34.3/ 29.5/ 31.9 & 26.0/ 20.0/ 27.0 & 21.0/ 24.0/ 18.0 & 25.2/ 25.5/ 25.2 & 20.8/ 24.2/ 23.6 \\
\bottomrule

\end{tabular}
}
\caption{Detailed subject-wise evaluation for \textsc{meta-llama/Llama-3.2-1B} on MILU across different languages. The results reported are X / Y / Z where X denote the 0-shot, Y denotes 1-shot and Z denotes 5-shot performance respectively.}
\end{table*}
\vspace*{\fill}
\endgroup

\begingroup
\setlength{\tabcolsep}{10pt} 
\renewcommand{\arraystretch}{1.3} 

\begin{table*}[]

\centering
\resizebox{\textwidth}{!}{
\begin{tabular}{lcccccccccccc}
\toprule
\textbf{Topic} & \multicolumn{1}{c}{\textit{\textbf{bn}}} & \multicolumn{1}{c}{\textit{\textbf{en}}} & \multicolumn{1}{c}{\textit{\textbf{gu}}} & \multicolumn{1}{c}{\textit{\textbf{hi}}} & \multicolumn{1}{c}{\textbf{\textit{kn}}} & \multicolumn{1}{c}{\textbf{\textit{ml}}} & \multicolumn{1}{c}{\textbf{\textit{mr}}} & \multicolumn{1}{c}{\textbf{\textit{or}}} & \multicolumn{1}{c}{\textbf{\textit{pa}}} & \multicolumn{1}{c}{\textbf{\textit{ta}}} & \multicolumn{1}{c}{\textbf{\textit{te}}} \\
\midrule
Agriculture & 31.0/ 25.0/ 24.0 & 26.8/ 26.8/ 25.6 & 31.3/ 29.3/ 24.2 & 27.4/ 24.8/ 23.3 & 31.0/ 24.0/ 25.0 & 30.3/ 24.2/ 23.2 & 29.0/ 23.0/ 28.0 & 27.0/ 30.0/ 24.0 & 27.0/ 22.0/ 23.0 & 22.0/ 23.0/ 26.0 & 18.0/ 22.0/ 26.0 \\
Anthropology & 26.0/ 33.0/ 30.0 & 26.0/ 27.0/ 23.0 & 29.0/ 30.0/ 25.0 & 33.3/ 37.4/ 25.2 & 29.0/ 26.0/ 31.0 & 24.0/ 25.0/ 21.0 & 29.0/ 27.0/ 24.0 & 31.0/ 29.0/ 20.0 & 21.0/ 21.0/ 18.0 & 21.0/ 25.0/ 30.0 & 27.0/ 27.0/ 26.0 \\
Architecture and Design & 19.0/ 18.0/ 19.0 & 23.0/ 23.0/ 19.0 & 23.0/ 28.0/ 20.0 & 36.0/ 25.0/ 22.0 & 26.0/ 23.0/ 21.0 & 28.0/ 30.0/ 28.0 & 32.0/ 28.0/ 21.0 & 26.0/ 27.0/ 25.0 & 32.0/ 27.0/ 27.0 & 31.0/ 25.0/ 27.0 & 21.0/ 26.0/ 22.0 \\
Arts and Culture & 32.7/ 26.7/ 31.5 & 32.4/ 25.8/ 27.4 & 22.0/ 25.0/ 25.0 & 25.4/ 28.1/ 28.1 & 23.9/ 22.7/ 21.5 & 29.0/ 26.0/ 26.0 & 22.4/ 23.5/ 25.1 & 23.0/ 27.0/ 23.0 & 27.0/ 28.0/ 29.0 & 29.7/ 27.9/ 27.9 & 22.6/ 25.6/ 29.1 \\
Astronomy and Astrophysics & 26.0/ 30.0/ 32.0 & 34.0/ 28.0/ 29.0 & 27.0/ 28.0/ 30.0 & 21.1/ 27.3/ 25.8 & 26.0/ 29.0/ 30.0 & 26.0/ 22.0/ 23.0 & 25.0/ 34.0/ 35.0 & 25.0/ 18.0/ 24.0 & 19.0/ 25.0/ 26.0 & 24.0/ 18.0/ 28.0 & 27.0/ 22.0/ 26.0 \\
Biology & 29.2/ 23.6/ 21.5 & 30.1/ 26.1/ 23.4 & 26.0/ 24.0/ 28.0 & 28.7/ 28.7/ 28.5 & 27.9/ 26.6/ 26.2 & 24.2/ 23.2/ 26.3 & 25.1/ 27.8/ 21.8 & 28.0/ 26.0/ 26.0 & 23.0/ 23.0/ 24.0 & 25.5/ 27.7/ 27.1 & 23.9/ 23.3/ 20.1 \\
\bottomrule

\end{tabular}
}
\caption{Detailed subject-wise evaluation for \textsc{meta-llama/Llama-3.2-1B-Instruct} on MILU across different languages. The results reported are X / Y / Z where X denote the 0-shot, Y denotes 1-shot and Z denotes 5-shot performance respectively.}
\end{table*}
\vspace*{\fill}
\endgroup

\begingroup
\setlength{\tabcolsep}{10pt} 
\renewcommand{\arraystretch}{1.3} 

\begin{table*}[]

\centering
\resizebox{\textwidth}{!}{
\begin{tabular}{lcccccccccccc}
\toprule
\textbf{Topic} & \multicolumn{1}{c}{\textit{\textbf{bn}}} & \multicolumn{1}{c}{\textit{\textbf{en}}} & \multicolumn{1}{c}{\textit{\textbf{gu}}} & \multicolumn{1}{c}{\textit{\textbf{hi}}} & \multicolumn{1}{c}{\textbf{\textit{kn}}} & \multicolumn{1}{c}{\textbf{\textit{ml}}} & \multicolumn{1}{c}{\textbf{\textit{mr}}} & \multicolumn{1}{c}{\textbf{\textit{or}}} & \multicolumn{1}{c}{\textbf{\textit{pa}}} & \multicolumn{1}{c}{\textbf{\textit{ta}}} & \multicolumn{1}{c}{\textbf{\textit{te}}} \\
\midrule
Agriculture & 25.0/ 21.0/ 20.0 & 42.9/ 48.1/ 49.3 & 27.3/ 26.3/ 27.3 & 26.2/ 28.0/ 27.4 & 16.0/ 20.0/ 25.0 & 24.2/ 28.3/ 25.2 & 20.0/ 27.0/ 24.0 & 25.0/ 32.0/ 27.0 & 23.0/ 29.0/ 34.0 & 22.0/ 30.0/ 26.0 & 28.0/ 24.0/ 28.0 \\
Anthropology & 33.0/ 33.0/ 36.0 & 48.0/ 49.0/ 58.0 & 31.0/ 34.0/ 36.0 & 29.3/ 32.3/ 38.4 & 27.0/ 34.0/ 34.0 & 30.0/ 38.0/ 39.0 & 32.0/ 41.0/ 38.0 & 23.0/ 23.0/ 20.0 & 23.0/ 28.0/ 27.0 & 33.0/ 29.0/ 27.0 & 28.0/ 31.0/ 28.0 \\
Architecture and Design & 27.0/ 18.0/ 21.0 & 37.0/ 47.0/ 46.0 & 28.0/ 23.0/ 19.0 & 31.0/ 30.0/ 27.0 & 33.0/ 31.0/ 24.0 & 29.0/ 24.0/ 24.0 & 30.0/ 20.0/ 20.0 & 26.0/ 22.0/ 22.0 & 36.0/ 35.0/ 36.0 & 26.0/ 21.0/ 24.0 & 24.0/ 25.0/ 22.0 \\
Arts and Culture & 20.6/ 25.4/ 21.8 & 48.4/ 55.8/ 57.8 & 29.0/ 23.0/ 27.0 & 28.3/ 28.5/ 29.1 & 26.4/ 27.0/ 28.2 & 22.0/ 31.0/ 28.0 & 25.7/ 26.8/ 23.5 & 19.0/ 20.0/ 21.0 & 25.0/ 27.0/ 33.0 & 32.1/ 32.7/ 33.3 & 22.1/ 26.6/ 24.6 \\
Astronomy and Astrophysics & 30.0/ 23.0/ 27.0 & 54.0/ 64.0/ 68.0 & 24.0/ 35.0/ 30.0 & 35.2/ 39.8/ 37.5 & 31.0/ 24.0/ 34.0 & 33.0/ 29.0/ 30.0 & 35.0/ 41.0/ 37.0 & 29.0/ 19.0/ 27.0 & 28.0/ 21.0/ 27.0 & 27.0/ 32.0/ 29.0 & 28.0/ 26.0/ 31.0 \\
Biology & 28.9/ 28.9/ 31.7 & 62.3/ 67.3/ 69.9 & 20.0/ 22.0/ 24.0 & 30.9/ 37.7/ 36.9 & 29.1/ 34.8/ 31.6 & 24.2/ 28.3/ 31.3 & 28.4/ 30.4/ 32.5 & 25.0/ 24.0/ 25.0 & 27.0/ 25.0/ 29.0 & 26.8/ 29.0/ 29.3 & 25.5/ 31.8/ 30.2 \\
\bottomrule

\end{tabular}
}
\caption{Detailed subject-wise evaluation for \textsc{google/gemma-2-2b} on MILU across different languages. The results reported are X / Y / Z where X denote the 0-shot, Y denotes 1-shot and Z denotes 5-shot performance respectively.}
\end{table*}
\vspace*{\fill}
\endgroup

\begingroup
\setlength{\tabcolsep}{10pt} 
\renewcommand{\arraystretch}{1.3} 

\begin{table*}[]

\centering
\resizebox{\textwidth}{!}{
\begin{tabular}{lcccccccccccc}
\toprule
\textbf{Topic} & \multicolumn{1}{c}{\textit{\textbf{bn}}} & \multicolumn{1}{c}{\textit{\textbf{en}}} & \multicolumn{1}{c}{\textit{\textbf{gu}}} & \multicolumn{1}{c}{\textit{\textbf{hi}}} & \multicolumn{1}{c}{\textbf{\textit{kn}}} & \multicolumn{1}{c}{\textbf{\textit{ml}}} & \multicolumn{1}{c}{\textbf{\textit{mr}}} & \multicolumn{1}{c}{\textbf{\textit{or}}} & \multicolumn{1}{c}{\textbf{\textit{pa}}} & \multicolumn{1}{c}{\textbf{\textit{ta}}} & \multicolumn{1}{c}{\textbf{\textit{te}}} \\
\midrule
Agriculture & 34.0/ 31.0/ 32.0 & 41.6/ 47.9/ 51.1 & 29.3/ 35.4/ 34.3 & 27.4/ 32.2/ 33.6 & 17.0/ 19.0/ 16.0 & 19.2/ 19.2/ 22.2 & 25.0/ 24.0/ 26.0 & 26.0/ 27.0/ 30.0 & 25.0/ 28.0/ 26.0 & 27.0/ 28.0/ 29.0 & 28.0/ 31.0/ 28.0 \\
Anthropology & 28.0/ 28.0/ 29.0 & 47.0/ 55.0/ 53.0 & 31.0/ 30.0/ 32.0 & 44.4/ 39.4/ 42.4 & 27.0/ 26.0/ 28.0 & 33.0/ 29.0/ 40.0 & 41.0/ 39.0/ 38.0 & 27.0/ 26.0/ 29.0 & 24.0/ 29.0/ 27.0 & 33.0/ 27.0/ 28.0 & 30.0/ 29.0/ 29.0 \\
Architecture and Design & 23.0/ 27.0/ 27.0 & 40.0/ 51.0/ 48.0 & 32.0/ 31.0/ 28.0 & 37.0/ 30.0/ 34.0 & 19.0/ 24.0/ 27.0 & 31.0/ 24.0/ 21.0 & 40.0/ 35.0/ 36.0 & 20.0/ 25.0/ 20.0 & 26.0/ 28.0/ 31.0 & 24.0/ 27.0/ 23.0 & 23.0/ 27.0/ 25.0 \\
Arts and Culture & 26.1/ 27.3/ 29.1 & 44.6/ 54.4/ 57.0 & 17.0/ 21.0/ 20.0 & 29.9/ 32.9/ 32.5 & 21.5/ 22.7/ 19.6 & 30.0/ 33.0/ 29.0 & 30.0/ 29.5/ 26.8 & 23.0/ 20.0/ 21.0 & 31.0/ 28.0/ 32.0 & 24.9/ 24.2/ 30.9 & 21.6/ 20.6/ 19.6 \\
Astronomy and Astrophysics & 31.0/ 28.0/ 32.0 & 66.0/ 71.0/ 76.0 & 23.0/ 25.0/ 26.0 & 35.2/ 38.3/ 42.2 & 29.0/ 27.0/ 28.0 & 30.0/ 33.0/ 33.0 & 37.0/ 37.0/ 38.0 & 25.0/ 20.0/ 22.0 & 24.0/ 23.0/ 28.0 & 33.0/ 32.0/ 31.0 & 32.0/ 29.0/ 31.0 \\
Biology & 31.0/ 29.9/ 26.8 & 60.1/ 71.3/ 71.5 & 25.0/ 22.0/ 21.0 & 34.9/ 40.5/ 40.5 & 25.8/ 26.2/ 27.9 & 29.3/ 34.3/ 34.3 & 31.0/ 31.3/ 32.5 & 26.0/ 27.0/ 31.0 & 36.0/ 30.0/ 33.0 & 32.8/ 31.5/ 31.2 & 30.5/ 33.0/ 30.5 \\
\bottomrule

\end{tabular}
}
\caption{Detailed subject-wise evaluation for \textsc{google/gemma-2-2b-it} on MILU across different languages. The results reported are X / Y / Z where X denote the 0-shot, Y denotes 1-shot and Z denotes 5-shot performance respectively.}
\end{table*}
\vspace*{\fill}
\endgroup

\begingroup
\setlength{\tabcolsep}{10pt} 
\renewcommand{\arraystretch}{1.3} 

\begin{table*}[]

\centering
\resizebox{\textwidth}{!}{
\begin{tabular}{lcccccccccccc}
\toprule
\textbf{Topic} & \multicolumn{1}{c}{\textit{\textbf{bn}}} & \multicolumn{1}{c}{\textit{\textbf{en}}} & \multicolumn{1}{c}{\textit{\textbf{gu}}} & \multicolumn{1}{c}{\textit{\textbf{hi}}} & \multicolumn{1}{c}{\textbf{\textit{kn}}} & \multicolumn{1}{c}{\textbf{\textit{ml}}} & \multicolumn{1}{c}{\textbf{\textit{mr}}} & \multicolumn{1}{c}{\textbf{\textit{or}}} & \multicolumn{1}{c}{\textbf{\textit{pa}}} & \multicolumn{1}{c}{\textbf{\textit{ta}}} & \multicolumn{1}{c}{\textbf{\textit{te}}} \\
\midrule
Agriculture & 23.0/ 23.0/ 28.0 & 27.2/ 33.6/ 33.2 & 31.3/ 27.3/ 24.2 & 28.4/ 28.2/ 32.8 & 19.0/ 26.0/ 29.0 & 25.2/ 28.3/ 32.3 & 27.0/ 22.0/ 27.0 & 31.0/ 26.0/ 25.0 & 22.0/ 27.0/ 26.0 & 31.0/ 23.0/ 29.0 & 32.0/ 24.0/ 27.0 \\
Anthropology & 25.0/ 19.0/ 27.0 & 32.0/ 29.0/ 34.0 & 27.0/ 31.0/ 35.0 & 25.2/ 30.3/ 34.3 & 28.0/ 20.0/ 29.0 & 24.0/ 13.0/ 24.0 & 22.0/ 22.0/ 24.0 & 29.0/ 33.0/ 30.0 & 21.0/ 30.0/ 27.0 & 22.0/ 19.0/ 25.0 & 25.0/ 30.0/ 28.0 \\
Architecture and Design & 24.0/ 20.0/ 26.0 & 31.0/ 34.0/ 30.0 & 26.0/ 30.0/ 26.0 & 31.0/ 27.0/ 30.0 & 30.0/ 29.0/ 26.0 & 24.0/ 18.0/ 19.0 & 27.0/ 28.0/ 25.0 & 29.0/ 27.0/ 22.0 & 29.0/ 33.0/ 30.0 & 21.0/ 25.0/ 31.0 & 22.0/ 28.0/ 31.0 \\
Arts and Culture & 26.7/ 24.9/ 29.1 & 29.2/ 36.8/ 37.0 & 30.0/ 26.0/ 30.0 & 22.7/ 24.6/ 27.5 & 21.5/ 28.8/ 28.2 & 22.0/ 23.0/ 24.0 & 25.1/ 25.7/ 25.1 & 21.0/ 39.0/ 32.0 & 28.0/ 31.0/ 30.0 & 29.1/ 25.4/ 26.1 & 19.6/ 22.6/ 24.1 \\
Astronomy and Astrophysics & 26.0/ 33.0/ 34.0 & 27.0/ 39.0/ 48.0 & 33.0/ 32.0/ 39.0 & 35.2/ 31.2/ 43.8 & 37.0/ 32.0/ 35.0 & 32.0/ 21.0/ 25.0 & 30.0/ 27.0/ 32.0 & 25.0/ 32.0/ 29.0 & 26.0/ 35.0/ 31.0 & 29.0/ 24.0/ 29.0 & 27.0/ 40.0/ 34.0 \\
Biology & 22.9/ 26.8/ 30.3 & 33.1/ 40.1/ 47.3 & 27.0/ 24.0/ 25.0 & 28.7/ 27.1/ 34.3 & 24.2/ 26.2/ 32.0 & 31.3/ 28.3/ 31.3 & 23.6/ 26.0/ 30.4 & 25.0/ 25.0/ 23.0 & 32.0/ 35.0/ 32.0 & 29.9/ 26.1/ 28.7 & 28.9/ 26.7/ 32.1 \\
\bottomrule

\end{tabular}
}
\caption{Detailed subject-wise evaluation for \textsc{sarvamai/sarvam-2b-v0.5} on MILU across different languages. The results reported are X / Y / Z where X denote the 0-shot, Y denotes 1-shot and Z denotes 5-shot performance respectively.}
\end{table*}
\vspace*{\fill}
\endgroup

\begingroup
\setlength{\tabcolsep}{10pt} 
\renewcommand{\arraystretch}{1.3} 

\begin{table*}[]

\centering
\resizebox{\textwidth}{!}{
\begin{tabular}{lcccccccccccc}
\toprule
\textbf{Topic} & \multicolumn{1}{c}{\textit{\textbf{bn}}} & \multicolumn{1}{c}{\textit{\textbf{en}}} & \multicolumn{1}{c}{\textit{\textbf{gu}}} & \multicolumn{1}{c}{\textit{\textbf{hi}}} & \multicolumn{1}{c}{\textbf{\textit{kn}}} & \multicolumn{1}{c}{\textbf{\textit{ml}}} & \multicolumn{1}{c}{\textbf{\textit{mr}}} & \multicolumn{1}{c}{\textbf{\textit{or}}} & \multicolumn{1}{c}{\textbf{\textit{pa}}} & \multicolumn{1}{c}{\textbf{\textit{ta}}} & \multicolumn{1}{c}{\textbf{\textit{te}}} \\
\midrule
Agriculture & 31.0/ 32.0/ 35.0 & 34.0/ 46.1/ 47.5 & 20.2/ 27.3/ 29.3 & 25.4/ 29.4/ 35.2 & 27.0/ 33.0/ 29.0 & 28.3/ 27.3/ 27.3 & 31.0/ 34.0/ 34.0 & 33.0/ 33.0/ 29.0 & 29.0/ 25.0/ 22.0 & 28.0/ 28.0/ 28.0 & 18.0/ 23.0/ 26.0 \\
Anthropology & 34.0/ 34.0/ 38.0 & 37.0/ 57.0/ 58.0 & 36.0/ 41.0/ 33.0 & 33.3/ 31.3/ 25.2 & 25.0/ 24.0/ 27.0 & 23.0/ 27.0/ 24.0 & 27.0/ 32.0/ 34.0 & 26.0/ 38.0/ 32.0 & 30.0/ 27.0/ 27.0 & 30.0/ 27.0/ 34.0 & 26.0/ 25.0/ 19.0 \\
Architecture and Design & 27.0/ 32.0/ 34.0 & 38.0/ 60.0/ 62.0 & 27.0/ 21.0/ 24.0 & 27.0/ 35.0/ 34.0 & 32.0/ 38.0/ 34.0 & 24.0/ 25.0/ 28.0 & 30.0/ 38.0/ 36.0 & 24.0/ 24.0/ 27.0 & 25.0/ 27.0/ 31.0 & 34.0/ 30.0/ 26.0 & 23.0/ 30.0/ 28.0 \\
Arts and Culture & 27.9/ 29.1/ 32.7 & 40.0/ 59.6/ 59.6 & 30.0/ 31.0/ 30.0 & 25.9/ 31.1/ 33.7 & 27.6/ 28.8/ 32.5 & 26.0/ 34.0/ 31.0 & 22.9/ 26.2/ 31.1 & 15.0/ 15.0/ 21.0 & 24.0/ 26.0/ 29.0 & 27.3/ 27.9/ 26.7 & 26.1/ 23.6/ 28.1 \\
Astronomy and Astrophysics & 29.0/ 28.0/ 34.0 & 43.0/ 56.0/ 69.0 & 29.0/ 39.0/ 35.0 & 31.2/ 33.6/ 34.4 & 37.0/ 39.0/ 35.0 & 31.0/ 34.0/ 37.0 & 33.0/ 34.0/ 38.0 & 24.0/ 28.0/ 22.0 & 30.0/ 34.0/ 31.0 & 27.0/ 35.0/ 37.0 & 34.0/ 38.0/ 34.0 \\
Biology & 25.7/ 28.5/ 33.1 & 45.5/ 62.5/ 66.7 & 25.0/ 28.0/ 27.0 & 32.7/ 34.9/ 39.5 & 33.2/ 34.4/ 33.6 & 28.3/ 28.3/ 25.2 & 30.8/ 32.8/ 35.8 & 28.0/ 35.0/ 33.0 & 29.0/ 34.0/ 35.0 & 28.3/ 33.4/ 29.9 & 27.4/ 31.4/ 32.4 \\
\bottomrule

\end{tabular}
}
\caption{Detailed subject-wise evaluation for \textsc{meta-llama/Llama-3.2-3B} on MILU across different languages. The results reported are X / Y / Z where X denote the 0-shot, Y denotes 1-shot and Z denotes 5-shot performance respectively.}
\end{table*}
\vspace*{\fill}
\endgroup

\begingroup
\setlength{\tabcolsep}{10pt} 
\renewcommand{\arraystretch}{1.3} 

\begin{table*}[]

\centering
\resizebox{\textwidth}{!}{
%
}
\caption{Detailed subject-wise evaluation for \textsc{meta-llama/Llama-3.2-3B-Instruct} on MILU across different languages. The results reported are X / Y / Z where X denote the 0-shot, Y denotes 1-shot and Z denotes 5-shot performance respectively.}
\end{table*}
\vspace*{\fill}
\endgroup

\begingroup
\setlength{\tabcolsep}{3pt} 
\renewcommand{\arraystretch}{1} 
\begin{table*}[t]
\centering
\resizebox{\textwidth}{!}{%
%
%
}
\caption{Detailed subject-wise evaluation for \textsc{nvidia/Nemotron-4-Mini-Hindi-4B-Base} on MILU across
different languages. The results reported are for 5-shot experiments.}
\end{table*}
\endgroup

\begingroup
\setlength{\tabcolsep}{10pt} 
\renewcommand{\arraystretch}{1.3} 

\begin{table*}[]

\centering
\resizebox{\textwidth}{!}{
%
}
\caption{Detailed subject-wise evaluation for \textsc{meta-llama/Llama-2-7b-hf} on MILU across different languages. The results reported are X / Y / Z where X denote the 0-shot, Y denotes 1-shot and Z denotes 5-shot performance respectively.}
\end{table*}
\vspace*{\fill}
\endgroup

\begingroup
\setlength{\tabcolsep}{10pt} 
\renewcommand{\arraystretch}{1.3} 

\begin{table*}[]

\centering
\resizebox{\textwidth}{!}{
%
}
\caption{Detailed subject-wise evaluation for \textsc{meta-llama/Llama-2-7b-chat-hf} on MILU across different languages. The results reported are X / Y / Z where X denote the 0-shot, Y denotes 1-shot and Z denotes 5-shot performance respectively.}
\end{table*}
\vspace*{\fill}
\endgroup

\begingroup
\setlength{\tabcolsep}{3pt} 
\renewcommand{\arraystretch}{1} 
\begin{table*}[t]
\centering
\resizebox{\textwidth}{!}{%
%
%
}
\caption{Detailed subject-wise evaluation for \textsc{neulab/Pangea-7B} on MILU across
different languages. The results reported are for 5-shot experiments.}
\end{table*}
\endgroup

\begingroup
\setlength{\tabcolsep}{10pt} 
\renewcommand{\arraystretch}{1.3} 

\begin{table*}[]

\centering
\resizebox{\textwidth}{!}{
%
}
\caption{Detailed subject-wise evaluation for \textsc{CohereForAI/aya-23-8B} on MILU across different languages. The results reported are X / Y / Z where X denote the 0-shot, Y denotes 1-shot and Z denotes 5-shot performance respectively.}
\end{table*}
\vspace*{\fill}
\endgroup

\begingroup
\setlength{\tabcolsep}{10pt} 
\renewcommand{\arraystretch}{1.3} 

\begin{table*}[]

\centering
\resizebox{\textwidth}{!}{
%
}
\caption{Detailed subject-wise evaluation for \textsc{meta-llama/Llama-3.1-8B} on MILU across different languages. The results reported are X / Y / Z where X denote the 0-shot, Y denotes 1-shot and Z denotes 5-shot performance respectively.}
\end{table*}
\vspace*{\fill}
\endgroup

\begingroup
\setlength{\tabcolsep}{10pt} 
\renewcommand{\arraystretch}{1.3} 

\begin{table*}[]

\centering
\resizebox{\textwidth}{!}{
%
}
\caption{Detailed subject-wise evaluation for \textsc{meta-llama/Llama-3.1-8B-Instruct} on MILU across different languages. The results reported are X / Y / Z where X denote the 0-shot, Y denotes 1-shot and Z denotes 5-shot performance respectively.}
\end{table*}
\vspace*{\fill}
\endgroup

\begingroup
\setlength{\tabcolsep}{10pt} 
\renewcommand{\arraystretch}{1.3} 

\begin{table*}[]

\centering
\resizebox{\textwidth}{!}{
%
}
\caption{Detailed subject-wise evaluation for \textsc{google/gemma-2-9b} on MILU across different languages. The results reported are X / Y / Z where X denote the 0-shot, Y denotes 1-shot and Z denotes 5-shot performance respectively.}
\end{table*}
\vspace*{\fill}
\endgroup

\begingroup
\setlength{\tabcolsep}{10pt} 
\renewcommand{\arraystretch}{1.3} 

\begin{table*}[]

\centering
\resizebox{\textwidth}{!}{
%
}
\caption{Detailed subject-wise evaluation for \textsc{google/gemma-2-9b-it} on MILU across different languages. The results reported are X / Y / Z where X denote the 0-shot, Y denotes 1-shot and Z denotes 5-shot performance respectively.}
\end{table*}
\vspace*{\fill}
\endgroup

\begingroup
\setlength{\tabcolsep}{10pt} 
\renewcommand{\arraystretch}{1.3} 

\begin{table*}[]

\centering
\resizebox{\textwidth}{!}{
%
}
\caption{Detailed subject-wise evaluation for \textsc{google/gemma-2-27b} on MILU across different languages. The results reported are X / Y / Z where X denote the 0-shot, Y denotes 1-shot and Z denotes 5-shot performance respectively.}
\end{table*}
\vspace*{\fill}
\endgroup

\begingroup
\setlength{\tabcolsep}{10pt} 
\renewcommand{\arraystretch}{1.3} 

\begin{table*}[]

\centering
\resizebox{\textwidth}{!}{
%
}
\caption{Detailed subject-wise evaluation for \textsc{google/gemma-2-27b-it} on MILU across different languages. The results reported are X / Y / Z where X denote the 0-shot, Y denotes 1-shot and Z denotes 5-shot performance respectively.}
\end{table*}
\vspace*{\fill}
\endgroup

\begingroup
\setlength{\tabcolsep}{10pt} 
\renewcommand{\arraystretch}{1.3} 

\begin{table*}[]

\centering
\resizebox{\textwidth}{!}{
%
}
\caption{Detailed subject-wise evaluation for \textsc{CohereForAI/aya-23-35B} on MILU across different languages. The results reported are X / Y / Z where X denote the 0-shot, Y denotes 1-shot and Z denotes 5-shot performance respectively.}
\end{table*}
\vspace*{\fill}
\endgroup

\begingroup
\setlength{\tabcolsep}{10pt} 
\renewcommand{\arraystretch}{1.3} 

\begin{table*}[]

\centering
\resizebox{\textwidth}{!}{
%
}
\caption{Detailed subject-wise evaluation for \textsc{meta-llama/Llama-3.1-70B} on MILU across different languages. The results reported are X / Y / Z where X denote the 0-shot, Y denotes 1-shot and Z denotes 5-shot performance respectively.}
\end{table*}
\vspace*{\fill}
\endgroup

\begingroup
\setlength{\tabcolsep}{3pt} 
\renewcommand{\arraystretch}{1} 
\begin{table*}[t]
\centering
\resizebox{\textwidth}{!}{%
%
}
\caption{0-shot subject-wise evaluation for \textsc{meta-llama/Llama-3.1-70B-Instruct} on MILU across different languages.}
\end{table*}

\begingroup
\setlength{\tabcolsep}{3pt} 
\renewcommand{\arraystretch}{1} 
\begin{table*}[t]
\centering
\resizebox{\textwidth}{!}{%
%
}
\caption{1-shot subject-wise evaluation for \textsc{meta-llama/Llama-3.1-70B-Instruct} on MILU across different languages.}
\end{table*}

\endgroup

\begingroup
\setlength{\tabcolsep}{3pt} 
\renewcommand{\arraystretch}{1} 
\begin{table*}[t]
\centering
\resizebox{\textwidth}{!}{%
%
}
\caption{Detailed subject-wise evaluation for \textsc{meta-llama/Llama-3.1-405B} on MILU across different languages. The results reported are for 0-shot experiments.}
\end{table*}
\vspace*{\fill}
\endgroup